\documentclass[]{CrossDistill_arxiv}

\usepackage{amsmath,amsfonts,bm}

\def\eqref#1{equation~\ref{#1}}

\def\1{\bm{1}}

\DeclareMathAlphabet{\mathsfit}{\encodingdefault}{\sfdefault}{m}{sl}
\SetMathAlphabet{\mathsfit}{bold}{\encodingdefault}{\sfdefault}{bx}{n}

\usepackage{hyperref}
\usepackage{url}

\DeclareMathOperator{\sg}{sg}

\usepackage{microtype}
\usepackage{graphicx}
\newsavebox{\arxivlogopreload}
\AtBeginDocument{\sbox{\arxivlogopreload}{%
\includegraphics[width=1pt]{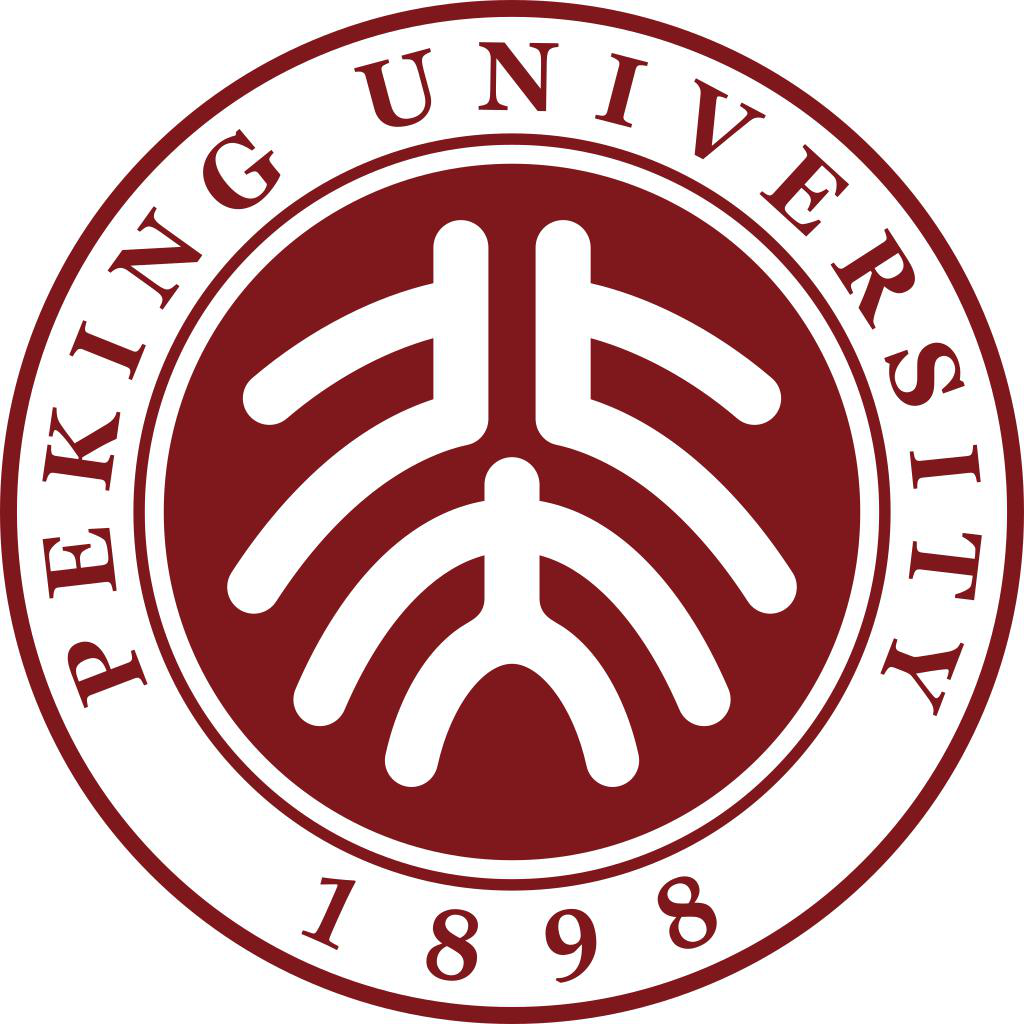}%
\includegraphics[width=1pt]{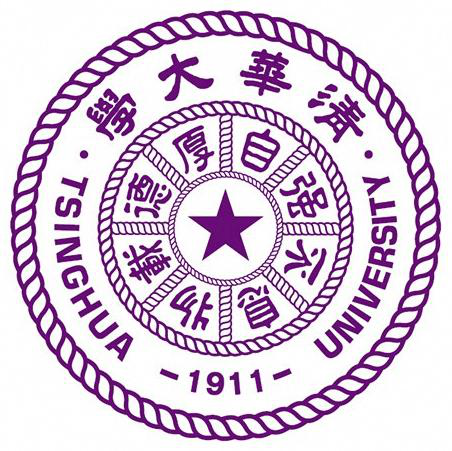}%
\includegraphics[width=1pt]{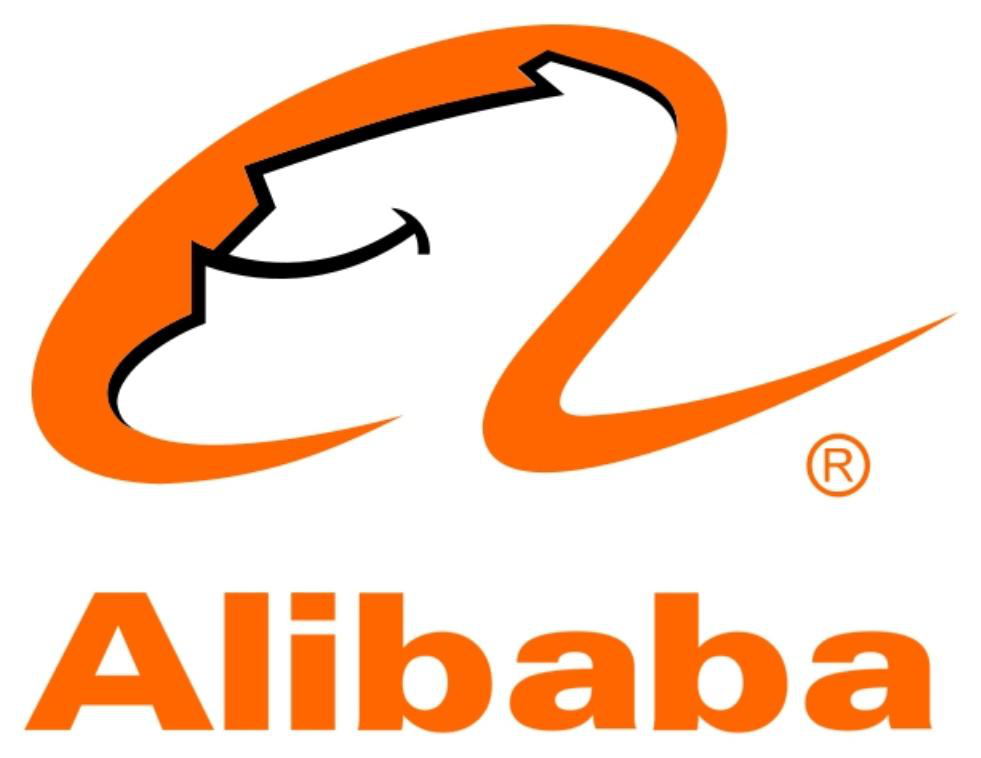}%
}}
\usepackage{subcaption}
\usepackage{xcolor}
\usepackage{colortbl}
\usepackage{amsfonts}       
\usepackage{nicefrac}       

\usepackage[table]{xcolor}         
\usepackage{tocvsec2}
\usepackage{titletoc}

\usepackage{xspace}
\usepackage{amsmath}
\usepackage{amsthm}
\usepackage{amssymb}

\usepackage{multicol}
\usepackage{multirow}
\usepackage{makecell}
\usepackage{enumitem}
\usepackage{wrapfig}
\usepackage{pifont}
\usepackage{graphicx,epstopdf}
\usepackage{needspace}

\usepackage{algorithm}     
\usepackage{algpseudocode} 
\usepackage{amsmath,amssymb} 

\usepackage{mathrsfs}

\usepackage{wasysym}

\usepackage{threeparttable, booktabs}
\usepackage{mathtools}
\mathtoolsset{showonlyrefs=true}
\usepackage{float}

\usepackage{tikz}

\definecolor{lightblue}{RGB}{230,246,253}  

\usepackage{tabularray}

\tcbuselibrary{skins,listings}
\newtcblisting{tikzexample}{
    sidebyside,
    center lower,
    bicolor,
    colbacklower=white,
    sharp corners,
    frame engine=empty
}

\definecolor{tridentRed}{HTML}{B9472F}
\definecolor{tridentOrange}{HTML}{E97845}
\definecolor{tridentLight}{HTML}{FFF0E9}
\definecolor{tridentHeader}{HTML}{FFF7F2}
\definecolor{tridentGray}{HTML}{666666}

\definecolor{lightblue}{RGB}{230,246,253}

\usepackage{amsmath}
\usepackage{amssymb}
\usepackage{mathtools}
\usepackage{amsthm}
\usepackage{longtable}

\usepackage{amsmath,amssymb,amsthm}

\makeatletter
\theoremstyle{plain}
\@ifundefined{theorem}{\newtheorem{theorem}{Theorem}[section]}{}
\@ifundefined{lemma}{}{}
\@ifundefined{proposition}{}{}
\@ifundefined{corollary}{}{}
\@ifundefined{assumption}{}{}
\@ifundefined{fact}{}{}
\theoremstyle{definition}
\@ifundefined{definition}{\newtheorem{definition}[theorem]{Definition}}{}
\@ifundefined{example}{}{}
\theoremstyle{remark}
\@ifundefined{remark}{}{}
\theoremstyle{plain}
\makeatother
\theoremstyle{plain}

\theoremstyle{definition}

\theoremstyle{remark}

\usepackage[disable,textsize=tiny]{todonotes}

\definecolor{sk}{RGB}{67,151,143}

\definecolor{lyx}{RGB}{221,160,221}

\definecolor{lhy}{RGB}{122,20,122}

\newtcolorbox{limbox}{
  enhanced,
  breakable,
  colback=orange!5,
  frame hidden,
  boxrule=0pt,
  borderline west={1.2pt}{0pt}{orange!55!black},
  sharp corners,
  left=6pt,
  right=2pt,
  top=3pt,
  bottom=3pt,
  before skip=5pt,
  after skip=5pt
}

\newtcolorbox{limbox_blue}{
  enhanced, breakable, colback=blue!5,
  frame hidden, boxrule=0pt,
  borderline west={1.2pt}{0pt}{blue!55!black},
  sharp corners, left=6pt, right=2pt, top=3pt, bottom=3pt,
  before skip=5pt, after skip=5pt
}

\usepackage{booktabs}

\definecolor{tridentRed}{RGB}{200, 16, 46}
\definecolor{tridentHeader}{RGB}{220, 230, 240}
\definecolor{tridentLight}{RGB}{245, 245, 245}
\newcommand{\best}[1]{\textbf{#1}}

\title{CrossDistill: Balancing Quality and Diversity via Trajectory-Level Hybrid Few-Step Distillation}

\author[13]{Yuxi Liu$^*$}
\author[23]{Haoyu Li$^*$}
\author[1]{Yixiang Cai}
\author[3]{Tengxu Sun$^\dagger$}
\author[1]{Zekun Zhang}
\author[3]{Baole Ai}
\author[3]{Ang Wang}
\author[3]{Jiamang Wang}
\author[3]{Lin Qu}
\author[1]{Kun Yuan$^\dagger$}
\author[2]{Kai Zhang$^\dagger$}

\affiliation[1]{Peking University, Melon Group}
\affiliation[2]{Tsinghua University}
\affiliation[3]{Alibaba Group}
\firstpagenotes{$^*$ Equal contribution.\par $^\dagger$ Corresponding author.}
\appto{\affiliationlist}{%
\par\vskip 1.5mm
{\affiliationfont\sffamily\bfseries
GitHub:
\href{https://github.com/AlibabaResearch/SparkDiffusion}
{\nolinkurl{https://github.com/AlibabaResearch/SparkDiffusion}}%
}%
}

\paperemail{%
\href{mailto:yuxiliu666@stu.pku.edu.cn}{yuxiliu666@stu.pku.edu.cn}
\quad
\href{mailto:hy-l24@mails.tsinghua.edu.cn}{hy-l24@mails.tsinghua.edu.cn}
\quad
\href{mailto:lingyan.sk@alibaba-inc.com}{lingyan.sk@alibaba-inc.com}

}

\date{\today}
\firstpagenotes{$^*$Equal contribution. $^\dagger$Corresponding author.}

\abstract{
Few-step distillation accelerates diffusion models but must balance diversity and fidelity: trajectory-based distillation preserves mode coverage, while distribution matching sharpens samples but can reduce diversity. We show that this tension can be exploited in a noise-regime-dependent way: high-noise steps largely determine global modes, whereas low-noise steps refine local details. We propose CrossDistill, a trajectory-level hybrid distillation framework that splits the sampling trajectory at a crossover point, applies a trajectory-preserving objective on the high-noise interval and a distribution-matching objective on the low-noise interval, and couples the two stages through the crossover state. In contrast to loss-level mixing, and complementarily to training-time two-stage recipes, CrossDistill explicitly assigns complementary objectives along the noise axis, so that global branching is preserved before local statistics are sharpened. CrossDistill is a noise-level scheduling policy: PCM and DMD are plug-in instantiations, while the noise partition, crossover coupling, and objective ordering are the key design elements. Experiments on text-to-video diffusion models and qualitative image-to-video results show that CrossDistill expands the few-step quality-diversity frontier, retaining seed-level variation while achieving competitive visual fidelity.
}

\begin{document}

\maketitle

\begin{figure}[htbp]          
    \centering                
\includegraphics[width=1.0\linewidth]{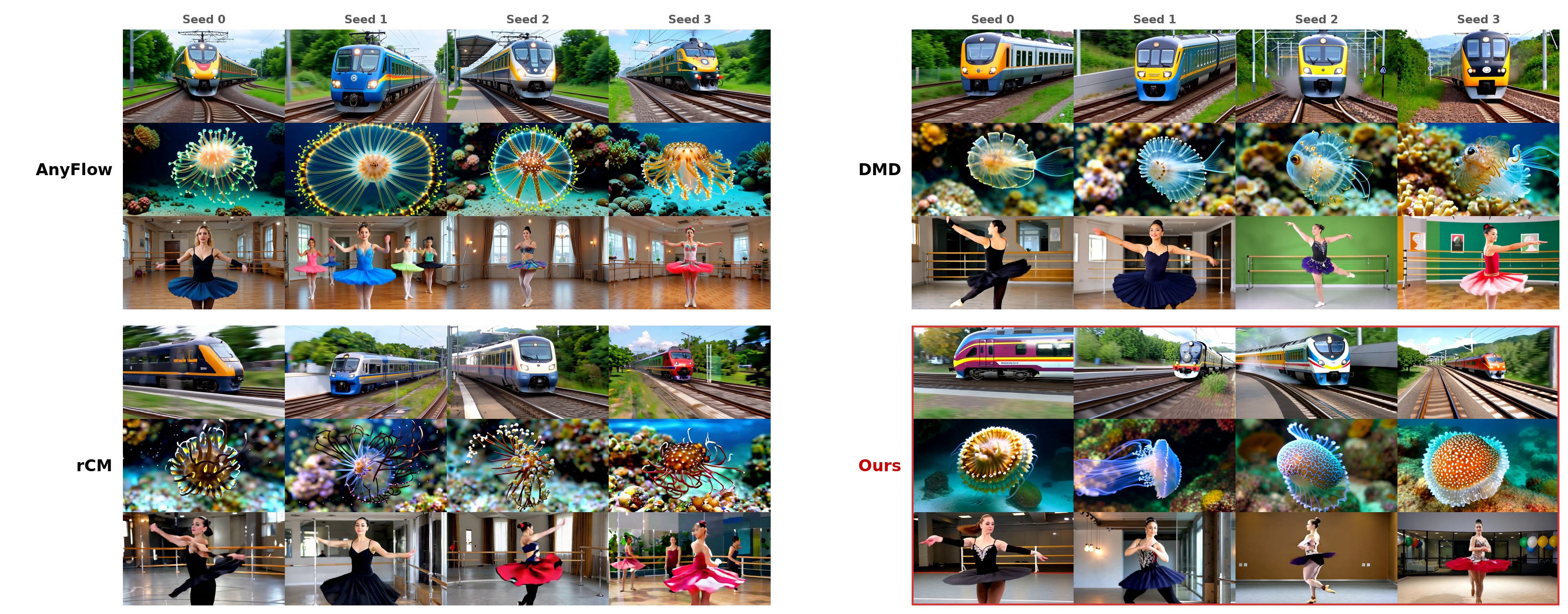}  
    \caption{\textbf{Diversity--quality trade-off.} For each of three prompts (rows) we sample four independent initial noise seeds (columns) and show the same frame per generated video. AnyFlow and DMD show reduced seed-level variation, while rCM is diverse but lower in visual quality. In these examples, CrossDistill improves the trade-off, maintaining competitive diversity and visual quality.}       
\label{fig:diversity_comparison_wide}      
\end{figure}



\section{Introduction}
\label{sec:intro}

Diffusion models are now a leading paradigm for high-fidelity image and video generation, but sampling remains expensive because it typically requires dozens of denoising steps. Few-step distillation trains a student model to reproduce the teacher's generative behavior with a small number of function evaluations (NFE). Existing distillation objectives fall into two complementary families. Trajectory-based distillation (TD) methods train the student to follow the teacher's probability-flow trajectory or to satisfy consistency relations across noise levels~\cite{salimans2022progressive,song2023consistency,luo2023lcm,kim2024consistency,pcm}. These objectives tend to preserve mode coverage and seed-level diversity, but with only a few steps the regressed trajectory may not land exactly on the data manifold, leading to blurred or over-smoothed samples. Distribution-matching (DM) methods instead align the student-generated distribution with the teacher distribution using score-based or adversarial objectives~\cite{yin2024one,yin2024improved,tdm,zhou2024score,luo2023diff}. They often produce sharper samples, but because they act primarily on marginals rather than trajectories, they can become more mode-seeking and may reduce stochastic diversity and video dynamics, as observed in our experiments.

This complementarity motivates a noise-regime view of few-step distillation. Diffusion sampling often proceeds in a coarse-to-fine manner: high-noise steps tend to select the global semantic mode, layout, and subject identity, while low-noise steps refine textures and local statistics. We therefore use noise level as an objective-assignment criterion: diversity is largely shaped at high noise, whereas fidelity is largely refined at low noise.
 If DM is applied too early, it may suppress trajectory branching before alternative modes are preserved; if TD is applied too late, it may fail to enforce the local constraints required for high visual fidelity.

Recent methods often combine multiple distillation objectives, but often leave the noise support of each objective implicit. For example, SenseFlow and rCM introduce auxiliary adversarial or trajectory regularization to stabilize distribution matching~\cite{senseflow, rcm}; DMDR enhances DMD by incorporating RL guidance~\cite{dmdr}; TDM combines trajectory awareness with distribution matching~\cite{tdm}.
Two-stage variants separate the objectives in training time rather than in noise level: \textit{From Structure to Detail}~\cite{Cheng2025From} trains with trajectory matching first and post-trains with distribution matching, while AnyFlow~\cite{gu2026anyflow} follows an off-policy flow-map stage with an on-policy post-training stage. In these recipes, however, the effective noise regime of each objective remains entangled with the choice of losses and training stages.

These observations motivate a noise-level assignment rule: assign TD to the high-noise interval, where mode structure is decided, and DM to the low-noise interval, where local fidelity is decided, and couple the two through a single crossover state. Loss-level mixing applies both objectives at every noise level, while two-stage sequencing separates objectives in training time but may still let each stage act on the whole noise range. We provide a conceptual discussion of this rule in Appendix~\ref{app:insight}; the empirical evidence is reported in Secs.~\ref{sec:evidence}--\ref{sec:ablation}.

We instantiate this rule as \textbf{CrossDistill}, a trajectory-level scheduling framework: a trajectory-preserving objective drives the high-noise interval $[\tau^\star,1]$ and a distribution-matching objective drives the low-noise interval $[0,\tau^\star]$, coupled through the crossover state. Concrete losses are plug-in choices of the form $\mathrm{Objective}^{\mathrm{H}}+\mathrm{Objective}^{\mathrm{L}}$; our default four-step student uses $\mathrm{PCM}^{\mathrm{H}}+\mathrm{DMD}^{\mathrm{L}}$, with the schedule rather than the loss pair as the design object. 
As shown in Fig.~\ref{fig:diversity_comparison_wide}, it retains seed-level variation while maintaining visual quality.

Our contributions are summarized as follows:
\begin{itemize}
        \item We formulate noise-level objective assignment as an explicit design axis in few-step diffusion distillation, and show that the objective ordering along the noise axis matters.

    \item We propose \textbf{CrossDistill}, a trajectory-level scheduling policy that couples a high-noise trajectory-preserving stage with a low-noise distribution-matching stage through a crossover state. The policy separates the scheduling decision from concrete TD/DM instantiations, e.g., $\mathrm{PCM}^{\mathrm{H}}$ or direct flow distillation ($\mathrm{DFD}^{\mathrm{H}}$; Sec.~\ref{sec:construction}) for the high-noise interval and $\mathrm{TDM}^{\mathrm{L}}/\mathrm{DMD}^{\mathrm{L}}$ for the low-noise interval.

        \item Experiments on text-to-video diffusion models, together with qualitative image-to-video results, show that, in our evaluated settings, CrossDistill improves the quality--diversity frontier relative to monolithic TD/DM baselines and loss-level hybrids.

\end{itemize}

\section{Hybrid Trajectory Distillation via a Crossover Point}
\label{sec:method}

\subsection{Preliminaries: Two Paradigms of Few-Step Distillation}
\label{sec:prelim}

\textbf{Setup.}
We consider a flow matching model where the forward process constructs a linear probability path between data and noise. 
Let $\mathbf{x}_0\sim p_{\mathrm{data}}$ and $\boldsymbol{\epsilon}\sim\mathcal{N}(\mathbf{0},\mathbf{I})$. The forward diffusion
$$
\mathbf{x}_t=(1-t)\mathbf{x}_0+t\boldsymbol{\epsilon},
\qquad t\in[0,1],
$$
induces a family of marginals $\{p_t\}$ with $p_0=p_{\mathrm{data}}$ and $p_1=\mathcal{N}(\mathbf{0},\mathbf{I})$. The teacher probability-flow ODE (PF-ODE) defines a flow map $\Psi_\phi^{\,t\to s}$ ($s\le t$) that deterministically transports samples of $p_t$ to $p_s$. 

\textbf{Trajectory-based distillation (TD).}
TD methods distill the teacher flow map or its induced trajectory structure: the student is required to progressively maintain the teacher's trajectory~\cite{salimans2022progressive}, or send states on a teacher trajectory to the same endpoint as the teacher~\cite{song2023consistency,kim2024consistency,geng2026mean}, so that a few student steps shadow the full ODE.  A representative instance is the phased consistency model (PCM)~\cite{pcm}, which partitions $[0,1]$ into sub-intervals, or phases, and enforces a self-consistency relation within each phase: all states on the same teacher trajectory segment are mapped to the same phase endpoint. Because such objectives are imposed along trajectories, the student inherits the teacher's mode coverage and diversity. Their limitation is geometric fidelity: with only a few steps, the regressed trajectories need not land exactly on the data manifold, which appears as blurred or over-smoothed samples.

\textbf{Distribution matching (DM).}
Distribution-matching methods instead align the student-induced marginals $q_s^\theta$ with the teacher marginals $p_s$ over a subset of noise levels, using the score difference $\nabla\log q_s^\theta-\nabla\log p_s$ estimated by a teacher denoiser and a fake model fit to student outputs, as in DMD~\cite{yin2024one} and its extensions~\cite{yin2024improved}. This pressure pushes the student onto the data manifold and yields sharp, high-fidelity samples. However, acting on marginals rather than individual trajectories, it is empirically prone to severe mode seeking and diversity degradation (Fig.~\ref{fig:prompt0_comparison}).

\subsection{The Design Principle: Preserve Global Branching at High Noise, Match Distribution at Low Noise}
\label{sec:principle}

\textbf{Coarse-to-fine responsibility as a scheduling criterion.}
The reverse diffusion process is often coarse to fine. We use this behavior as an objective-assignment criterion: high-noise states are associated with global mode commitment, whereas low-noise states are associated with local detail refinement~\cite{biroli2024dynamical,sclocchi2025phase}.
 Once a trajectory has largely committed to a mode at high noise, later low-noise steps can refine that mode but typically cannot fully recover discarded alternatives, especially under deterministic few-step sampling.

\begin{figure}[t]
    \centering
    \includegraphics[width=0.75\textwidth]{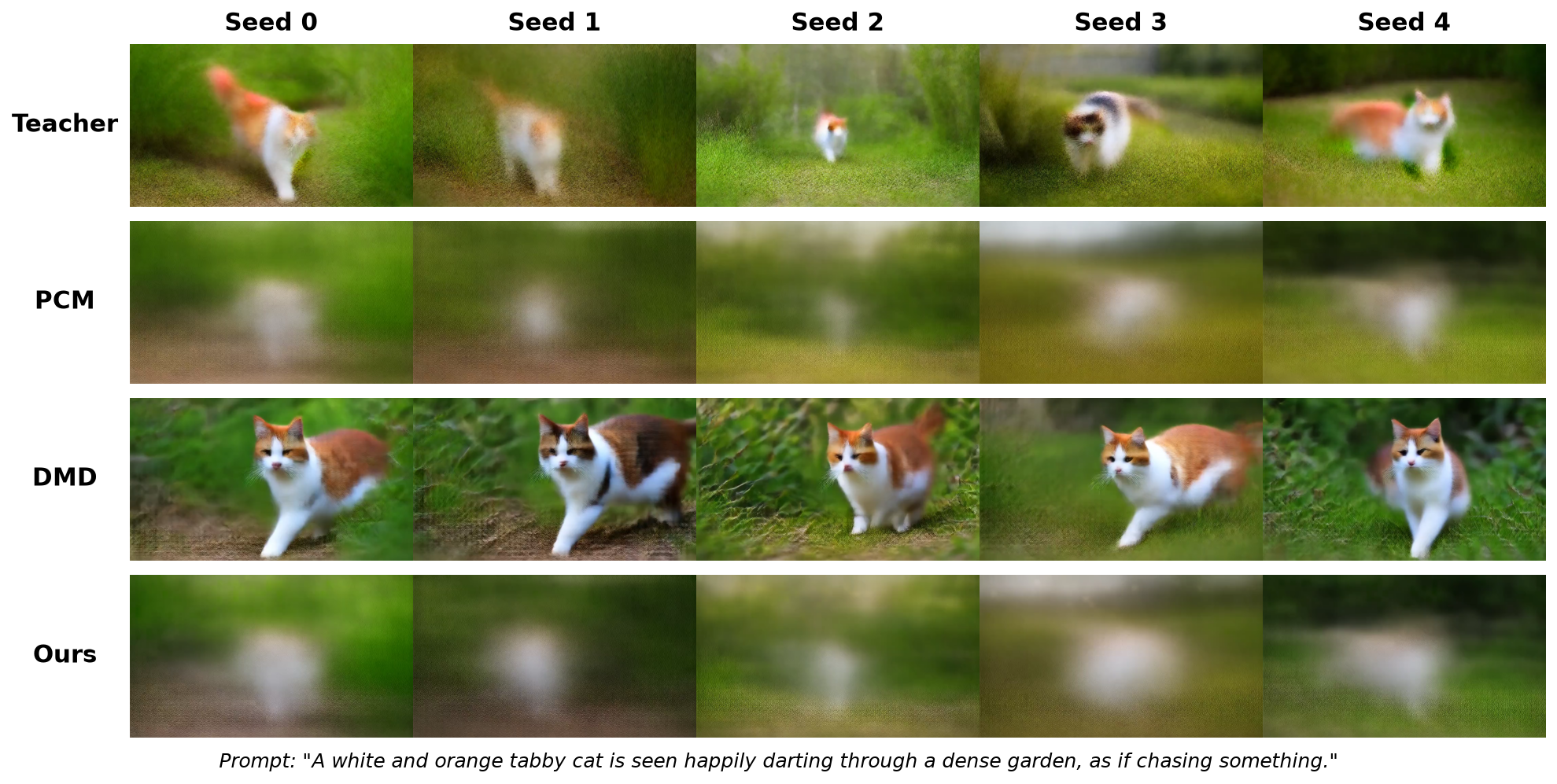}
    \caption{Intermediate denoising states in the high-noise regime ($t\approx0.94$) for the teacher and four-step distilled models, generated from the same text prompt and different initial noises. DMD already maps distinct seeds to similar states, whereas the teacher, PCM, and CrossDistill retain seed-level variation.}
    \label{fig:prompt0_comparison}
\end{figure}

Fig.~\ref{fig:prompt0_comparison} illustrates this behavior in the distillation regime. At high noise levels (around $t=0.94$), the purely DM-distilled model (DMD) already maps different initial noises to similar states, whereas the teacher, PCM, and CrossDistill retain variation. This motivates using TD in the noise-dominated regime to preserve stochastic diversity and DM at low noise to refine local details.

\textbf{Conceptual bookkeeping.}
Appendix~\ref{app:insight} distills the above responsibility split into a two-price ledger:
$w_D(t)$, the diversity cost of using DM instead of TD at level $t$, and $w_Q(t)$, the
quality cost of the opposite choice. The ledger is a conceptual aid rather than a training-dynamics model or a formal guarantee:
 its coarse-to-fine price ordering restates the responsibility
split above, is consistent with prior observations of coarse-to-fine
denoising~\cite{ho2020denoising,meng2021sdedit,choi2021ilvr,hertz2022prompt} and with our
diagnostics (Fig.~\ref{fig:prompt0_comparison}, Table~\ref{tab:reversed_assignment}), and
is checked where checkable (Fig.~\ref{fig:qd_pareto_mix}). Within this simplified accounting, per-level loss mixing corresponds to averaging two different targets, uniform blends correspond to sweeping a chord between two endpoints, and the reversed order would pay the more expensive prices. The empirical case for the rule rests on Secs.~\ref{sec:evidence}--\ref{sec:ablation}, not on the ledger.

\begin{figure}[H]
    \centering
    \includegraphics[width=0.85\textwidth]{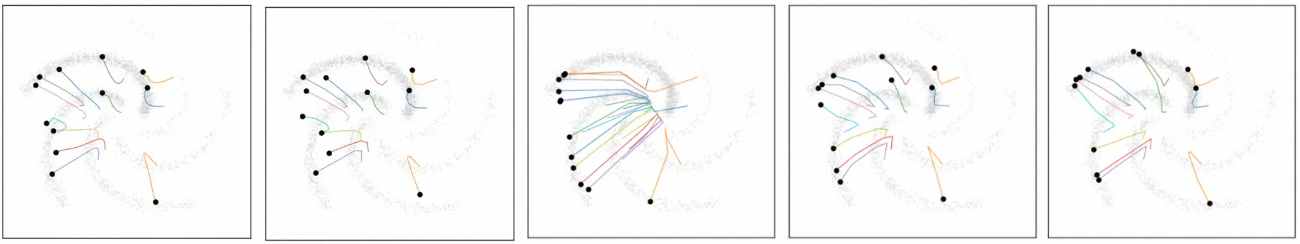}
    \caption{
    2D toy manifold distillation setting.
    From left to right: original teacher, trajectory-based distillation (TD), distribution matching (DM),
    loss-level mixture of TD and DM, and trajectory-level hybrid distillation (CrossDistill; ours).
    All models are trained on the same gray data manifold, and each student is trained to convergence.
    Colored curves denote denoising trajectories of 12 seeds, and black dots denote final generated samples.
    }
    \label{fig:2D_toy_case}
\end{figure}

\textbf{Toy manifold illustration.}
Fig.~\ref{fig:2D_toy_case} shows a similar tendency on a 2D toy manifold.
The TD student preserves the teacher's trajectories and covers the data distribution, but does not always reach the data manifold; the DM student fits the manifold but, lacking trajectory constraints, collapses to a subset of modes; the loss-level mixture compromises both. In this toy setting, CrossDistill improves both manifold fitting and mode coverage, providing a controlled illustration of trajectory-level scheduling rather than global loss-level mixing.

\begin{limbox}
\noindent\textbf{Design principle.} \textit{In our framework, we use trajectory-based distillation at high noise to preserve global mode structure, and distribution matching at low noise to refine local statistics. We combine them through a single trajectory-level crossover. The resulting rule is a noise-level scheduling policy with disjoint supports; concrete TD and DM losses are plug-in instantiations of this policy.}
\footnote{Appendix~\ref{app:insight} provides an optional conceptual discussion of this trade-off. It is intended only to clarify the intuition.
}
\end{limbox}

\textbf{Why not reverse the assignment?}
The opposite assignment, DM on $\mathcal{H}$ and TD on $\mathcal{L}$, would invert the coarse-to-fine responsibilities. At high noise, distribution matching acts on marginal statistics before alternative global modes have been stabilized, and therefore may pull different seeds toward common high-density modes. Once this seed-level branching is removed, a low-noise TD objective may be unable to recover discarded global alternatives, because it can only refine the trajectory that has already been selected. Conversely, low-noise fidelity depends on matching local data statistics, which TD does not enforce explicitly. Therefore, the reversed schedule is expected to sacrifice diversity without providing a compensating mechanism for local fidelity. To test this prediction, we implement a reversed control in Sec.~\ref{sec:ablation}, where the same crossover partition and NFE budget are kept but the noise supports of the TD and DM objectives are swapped. As shown in Table~\ref{tab:reversed_assignment}, the reversed variant underperforms the proposed ordering in both diversity and quality in our setting, supporting the coarse-to-fine assignment.

\subsection{CrossDistill via a Crossover Point}
\label{sec:construction}

\begin{definition}[Crossover partition]
Fix a crossover level $\tau^\star\in(0,1)$ and define the high-noise interval
$\mathcal{H}=[\tau^\star,1]$ and the low-noise interval
$\mathcal{L}=[0,\tau^\star]$. The shared boundary $\{\tau^\star\}$ has measure zero and can be assigned to either stage.
\end{definition}

\begin{definition}[Relay generator]
The student generator $G_\theta$ consists of a high-noise TD map
$f_\theta:\mathbb{R}^d\times\mathcal{H}\to\mathbb{R}^d$ and a low-noise sub-sampler $g_\theta$. In the default four-step setting, $f_\theta$ corresponds to one PCM step on $\mathcal{H}$, and $g_\theta$ corresponds to a three-step DMD sub-sampler on $\mathcal{L}$. Abstractly, the two-stage sampler is
\begin{equation}
  \boldsymbol{\epsilon}\sim p_{1}\ \longmapsto\
  \mathbf{x}_{\tau^\star}:=f_\theta(\boldsymbol{\epsilon},1)\ \longmapsto\
  \hat{\mathbf{x}}_0:=g_\theta(\mathbf{x}_{\tau^\star}).
  \label{eq:relay}
\end{equation}
\end{definition}

The policy specifies the noise supports and the relay interface;
 the objectives
below describe our default instantiation. In principle, other trajectory-preserving objectives may
replace Eq.~\eqref{eq:pcm} on $\mathcal{H}$, and other distribution-matching
objectives may replace Eq.~\eqref{eq:dmd} on $\mathcal{L}$; Sec.~\ref{sec:ablation} varies both within a limited set of choices.

\textbf{High-noise stage: trajectory-based distillation (TD).}
The map $f_\theta$ is constrained to be invariant along the teacher PF-ODE on $\mathcal{H}$ and anchored at $\tau^\star$, i.e.
$f_\theta(\mathbf{x}_t,t)=f_\theta(\Psi_\phi^{\,t\to t'}(\mathbf{x}_t),t')$ for all $t'\in\mathcal{H}$ with $t'\le t$, and $f_\theta(\cdot,\tau^\star)=\mathrm{id}$. This is enforced by consistency loss 
\begin{equation}
  \mathcal{L}_{\mathrm{PCM}}(\theta)=
  \mathbb{E}_{t\sim\mathcal{U}(\mathcal{H}),\,\mathbf{x}_t\sim p_t}
  \big\|\,f_\theta(\mathbf{x}_t,t)
        -\mathrm{sg} \big[f_\theta(\Psi_\phi^{\,t\to t'}(\mathbf{x}_t),t')\big]\,\big\|_2^2,
  \label{eq:pcm}
\end{equation}

where $t'$ denotes the next node of the student's fixed discretization of $\mathcal{H}$ toward $\tau^\star$; in the default one-step high-noise setting, $t'=\tau^\star$. Its intended fixed point is the teacher-induced crossover state at $\tau^\star$.

For the direct flow distillation (DFD) instantiation, $f_\theta$ directly regresses to the teacher-induced crossover state. Let
\begin{equation}
\mathbf{x}_{\tau^\star}^{\phi}:=\Psi_\phi^{\,1\to\tau^\star}(\mathbf{x}_1)
\end{equation}
be the state obtained by flowing $\mathbf{x}_1$ with the teacher PF-ODE or its multi-step discrete solver to $\tau^\star$. The DFD objective is formulated as a simple MSE loss:
\begin{equation}
\mathcal{L}_{\mathrm{DFD}}(\theta)=
\mathbb{E}_{\mathbf{x}_1\sim p_1}
\big\|f_\theta(\mathbf{x}_1,1)-\sg\big[\mathbf{x}_{\tau^\star}^{\phi}\big]\big\|_2^2.
\label{eq:dfd_high}
\end{equation}
Under a budget of four function evaluations (NFE$=4$), this corresponds to a single high-noise step from $t=1$ to $t=\tau^\star$. We denote these high-noise TD instantiations by $\mathrm{DFD}^{\mathrm{H}}$ and $\mathrm{PCM}^{\mathrm{H}}$, respectively.

\textbf{Low-noise stage: distribution matching (DM).} Let $\hat{\mathbf{x}}_0=g_\theta(\mathbf{x}_{\tau^\star})$ denote the output of the low-noise sub-sampler. For $s\in\mathcal{L}$, sample $\boldsymbol{\epsilon}\sim\mathcal{N}(0,\mathbf I)$ and form the noised state
\begin{equation}
\mathbf{x}_s=(1-s)\hat{\mathbf{x}}_0+s\boldsymbol{\epsilon}.
\end{equation}
We parameterize the teacher and fake denoiser as $F_\phi$ and $F_\psi$. The low-noise DMD objective is
\begin{equation}
  \mathcal{L}_{\mathrm{DMD}}(\theta)=
  \mathbb{E}_{s\sim\mathcal{U}(\mathcal{L}),\,\boldsymbol{\epsilon}\sim\mathcal{N}(0,\mathbf I)}
  \left\|
  \hat{\mathbf{x}}_0
  -
  \sg\!\left(
  \hat{\mathbf{x}}_0
  -
  \frac{
  F_\psi(\mathbf{x}_s,s,c)-F_\phi(\mathbf{x}_s,s,c)
  }{
  \mathrm{mean}
    |\hat{\mathbf{x}}_0-F_\phi(\mathbf{x}_s,s,c)|
  }
  \right)
  \right\|_2^2.
  \label{eq:dmd}
\end{equation}
The difference between the fake and teacher clean predictions approximates the difference between the student-induced and teacher score fields. Minimizing $\mathcal{L}_{\mathrm{DMD}}$ encourages the student-induced marginals to approach the teacher marginals on $\mathcal{L}$.

The fake model is trained on stopped student-generated samples
\begin{equation}
\mathcal{L}_{\mathrm{fake}}(\psi)
=
\mathbb{E}_{\hat{\mathbf{x}}_0,\,t\sim p_{\mathrm{fake}}}
\left\|
F_\psi(\mathbf{x}_t,t,c)
-\hat{\mathbf{x}}_0
\right\|_2^2,
\label{eq:fake_fm}
\end{equation}
where $\mathbf{x}_t=(1-t)\hat{\mathbf{x}}_0+t\boldsymbol{\epsilon}$, and $p_{\mathrm{fake}}$ is the noise-level distribution used for fake training.


\textbf{Training and gradient flow.}
The two generator objectives have disjoint temporal support:
$\operatorname{supp}_t\mathcal{L}_{\mathrm{PCM}}=\mathcal{H}$ and
$\operatorname{supp}_t\mathcal{L}_{\mathrm{DMD}}=\mathcal{L}$,
and are coupled through the shared crossover state $\mathbf{x}_{\tau^\star}$ in~\eqref{eq:relay}. In implementation, we optimize
\begin{equation}
\mathcal{L}_{G}
=
\lambda_{\mathrm{H}}\,\mathcal{L}_{\mathcal{O}_{\mathrm{H}}}
+
\lambda_{\mathrm{L}}\,\mathcal{L}_{\mathcal{O}_{\mathrm{L}}},
\qquad
\min_{\psi}\ \mathcal{L}_{\mathrm{fake}},
\label{eq:train}
\end{equation}
where the default instantiation uses
$(\mathcal{O}_{\mathrm{H}},\mathcal{O}_{\mathrm{L}})=(\mathrm{PCM},\mathrm{DMD})$.
Because $\mathcal{L}_{\mathrm{PCM}}$ is sampled only on $\mathcal{H}$ and $\mathcal{L}_{\mathrm{DMD}}$ is sampled only on $\mathcal{L}$, this aggregation does not impose competing TD and DM targets at the same noise level.
 In a training rollout, the high-noise relay first produces $\mathbf{x}_{\tau^\star}$, and the low-noise sub-sampler then produces the final clean prediction $\hat{\mathbf{x}}_0$. The DMD loss is applied to this low-noise output, while the fake model is trained on stopped rollouts. By default, the DMD gradient is stopped at the crossover state and is not back-propagated through the PCM relay. Thus the high-noise relay is directly supervised by the TD objective, while the low-noise distribution-matching pressure is intended to refine local statistics without directly back-propagating into the high-noise mode-selection relay.
 The low-noise stage is nevertheless trained on crossover states produced by the current high-noise relay, so the two stages remain coupled through the empirical distribution of $\mathbf{x}_{\tau^\star}$.

\textbf{Default step allocation under a fixed NFE budget.}
In the default four-step student, one function evaluation goes to the high-noise relay and three to the low-noise DM sampler. This follows from the role of $\tau^\star$: around $0.94$ the sample has already crossed the main mode-commitment region (Fig.~\ref{fig:crossover_grid_visual}), so the high-noise stage needs only to preserve the selected branch and produce a reliable crossover state. Extra high-noise steps would mainly refine an already committed structure, giving diminishing diversity returns while consuming budget that is more valuable for low-noise distribution matching. Here $g_\theta$ consists of three student velocity evaluations on a fixed low-noise timestep grid, and DMD supervises the final output. Under the NFE$=2$ budget, the same principle gives one TD step and one DM step.

\textbf{Instantiations of the scheduling policy.}
We view CrossDistill as a scheduling policy whose effect is not tied to a single fixed loss pair.
 We write an instantiation as
$
\text{CrossDistill}(\mathcal{O}_{\mathrm{H}},\mathcal{O}_{\mathrm{L}}),
$
where $\mathcal{O}_{\mathrm{H}}$ is applied on $\mathcal{H}$ and $\mathcal{O}_{\mathrm{L}}$ is applied on $\mathcal{L}$. In this paper, $\mathcal{O}_{\mathrm{H}}\in\{\mathrm{DFD}^{\mathrm{H}},\mathrm{PCM}^{\mathrm{H}}\}$ and $\mathcal{O}_{\mathrm{L}}\in\{\mathrm{TDM}^{\mathrm{L}},\mathrm{DMD}^{\mathrm{L}}\}$. Here $\mathrm{DFD}^{\mathrm{H}}$ is the direct regression objective of Eq.~\eqref{eq:dfd_high}, $\mathrm{PCM}^{\mathrm{H}}$ is the high-noise consistency relay of Eq.~\eqref{eq:pcm}, $\mathrm{DMD}^{\mathrm{L}}$ is the low-noise distribution-matching objective of Eq.~\eqref{eq:dmd}, and $\mathrm{TDM}^{\mathrm{L}}$ replaces it with the TDM objective~\cite{tdm} on the same low-noise interval. The default instantiation used in Table~\ref{tab:vbench_main} is
$
\text{CrossDistill}(\mathrm{PCM}^{\mathrm{H}},\mathrm{DMD}^{\mathrm{L}}),
$
which we abbreviate as CrossDistill when no ambiguity arises.

\subsection{Trajectory-Level vs.\ Loss-Level Mixing}
\label{sec:evidence}

\setlength{\columnsep}{8pt}
\setlength{\intextsep}{2pt}

\begin{wrapfigure}[11]{r}{0.65\textwidth}
    \vspace{0pt}
    \centering

    \begin{subfigure}[t]{0.49\linewidth}
        \centering
        \includegraphics[
            width=\linewidth,
            trim=2mm 1mm 2mm 1mm,
            clip
        ]{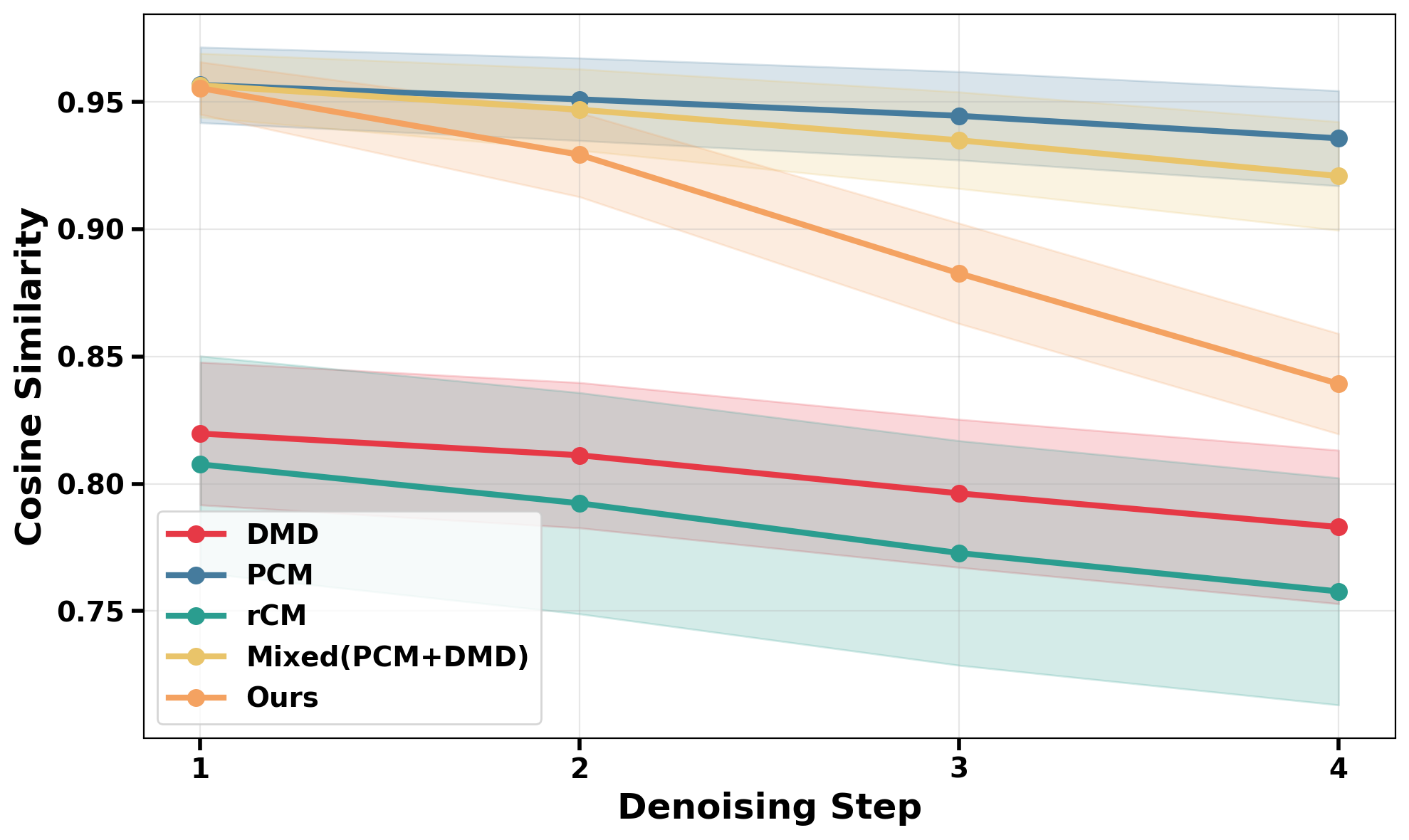}
        \caption{Cosine similarity.}
        \label{fig:cosine}
    \end{subfigure}\hfill%
    \begin{subfigure}[t]{0.49\linewidth}
        \centering
        \includegraphics[
            width=\linewidth,
            trim=2mm 1mm 2mm 1mm,
            clip
        ]{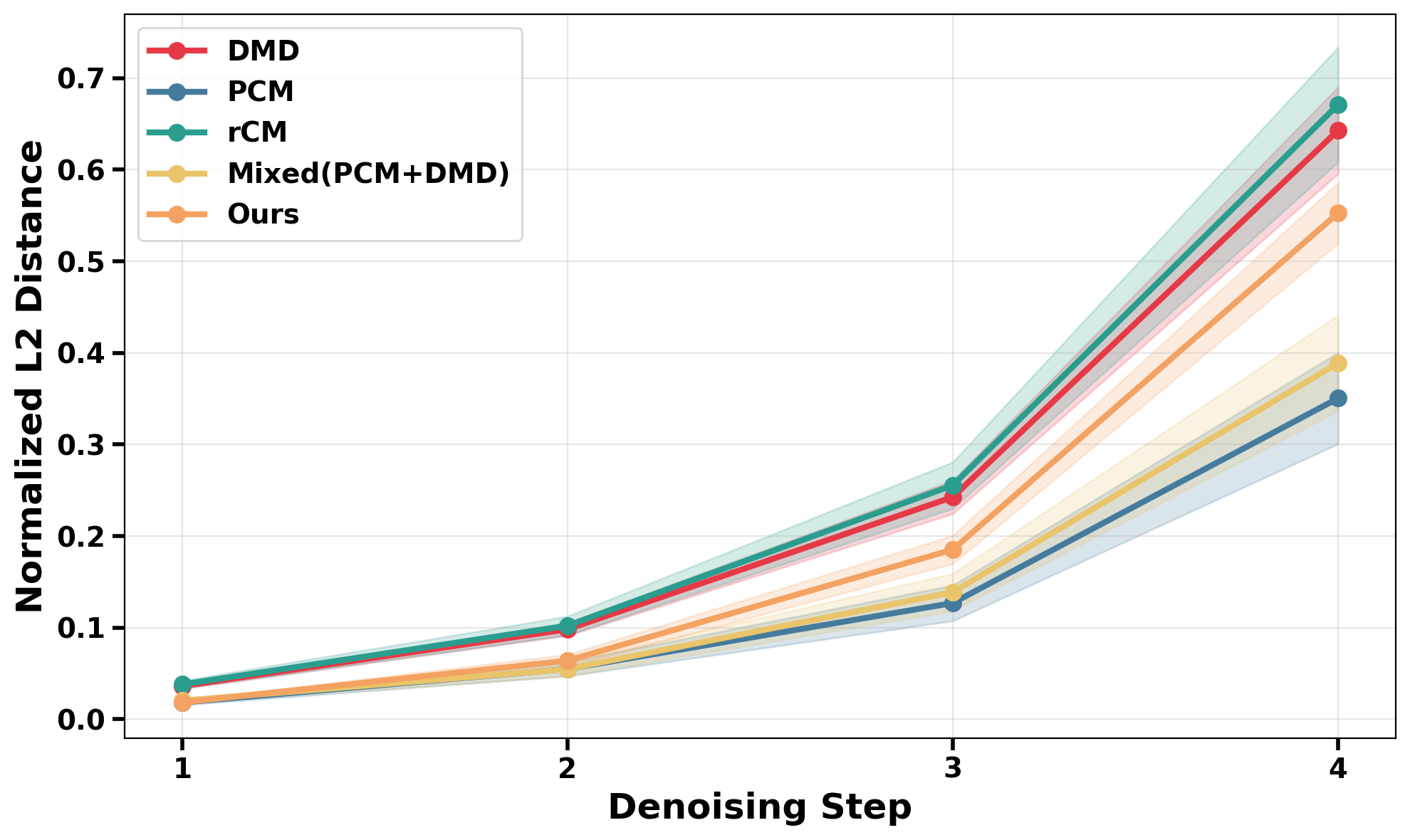}
        \caption{$\ell_2$ distance.}
        \label{fig:l2}
    \end{subfigure}

    \vspace{-2pt}

    \captionsetup{font=footnotesize,skip=2pt}
    \caption{Trajectory alignment metrics between student and teacher trajectories during denoising.}
    \label{fig:trajectory_alignment}
\end{wrapfigure}

Fig.~\ref{fig:trajectory_alignment} measures cosine similarity and $\ell_2$ distance between the student's intermediate states and the teacher's PF-ODE states at the same noise level, starting from the same initial noise. CrossDistill follows PCM in the high-noise regime and DMD in the low-noise regime, whereas the tuned loss-level mixture (Sec.~\ref{sec:main}) and rCM remain intermediate across noise levels: less teacher-like than PCM at high noise and less teacher-like than DMD at low noise. Trajectory-level scheduling therefore appears to retain each paradigm's strength in its effective noise regime.

\textbf{Relation to two-stage sequential training.}
A third combination axis is training-time sequencing: \textit{From Structure to Detail}~\cite{Cheng2025From} applies TD first and DM later, while AnyFlow~\cite{gu2026anyflow} sequences an off-policy MeanFlow stage with an on-policy DMD stage. Sequencing avoids simultaneous conflict between objectives, but without an explicit noise-axis partition the later stage can still act on high-noise mode-selection steps. CrossDistill instead provides an explicit noise-axis partition (Sec.~\ref{sec:construction}); this choice is orthogonal to sequencing, since a two-stage curriculum could equally be applied within each interval.
Empirically, our comparisons cover the cells of this design space: the tuned mixture (Fig.~\ref{fig:qd_pareto_mix}) shares CrossDistill's losses, training protocol, and budget, isolating the noise partition under joint training; AnyFlow (Table~\ref{tab:vbench_main}) represents the sequential cell under identical evaluation; and the reversed control (Table~\ref{tab:reversed_assignment}) isolates the ordering.

\section{Experiments}
\label{sec:exp}

\subsection{Experimental Setup}
\label{sec:setup}

\textbf{Models and data.}
We instantiate the CrossDistill scheduling policy on pre-trained Wan2.1 text-to-video diffusion transformers and distill them into few-step students. Unless otherwise stated, the reported model is the default instantiation $\mathrm{PCM}^{\mathrm{H}}+\mathrm{DMD}^{\mathrm{L}}$: one high-noise PCM step on $[\tau^\star,1]$ and three low-noise DMD steps on $[0,\tau^\star]$, with $\tau^\star=0.94$. For image-to-video, we use Wan2.1 I2V at 720P with the same four-step protocol and report qualitative comparisons.

\textbf{Baselines.}
We compare with:
(i) Base teacher model, the full-step reference;
(ii) \textbf{PCM}~\cite{pcm} and (iii) \textbf{DMD}~\cite{yin2024one}, monolithic TD and DM baselines;
(iv) \textbf{rCM}~\cite{rcm};
(v) \textbf{AnyFlow}~\cite{gu2026anyflow}, an any-step student distilled through a two-stage process; and
(vi) \textbf{loss-level mixture}, a single student trained with $\lambda_{\mathrm{PCM}}\mathcal{L}_{\mathrm{PCM}}+\lambda_{\mathrm{DMD}}\mathcal{L}_{\mathrm{DMD}}$ applied over the full noise range.
Our method is denoted \textbf{CrossDistill}; alternative instantiations of the same policy are studied in Sec.~\ref{sec:ablation}. Since the loss-level mixture contains multiple weight settings, we report its quality--diversity frontier in Fig.~\ref{fig:qd_pareto_mix} and qualitative comparisons among the mixed, reversed, and proposed schedules in Fig.~\ref{fig:ours_mixed_reversed}.
For PCM, DMD, rCM, and AnyFlow we evaluate the officially released checkpoints distilled from the same Wan2.1 teachers, so that each baseline is the strongest published realization of its method; the loss-level mixture and all CrossDistill variants are trained by us under an identical protocol (same teacher, data, training iterations, optimizer, and critic protocol; Table~\ref{tab:training_hyperparameters}). All methods are evaluated on the same prompts, initial noise seeds, resolution, and NFE budget.

\textbf{Metrics.}
We report VBench quality, semantic, and total scores~\cite{vbench} following the official protocol. To evaluate stochastic diversity, we measure the input-dependent variation directly in the embedding spaces of two frozen video encoders, V-JEPA~2~\cite{assran2025v} and VideoMAE~V2~\cite{videomaev2}, following PDD\cite{pdd}. Specifically, each generated video is encoded into a global embedding, and the diversity score is computed as the average pairwise distance (both cosine and $\ell_2$) among $N$ videos generated from the same prompt with different initial noises. The V-JEPA~2 cosine distance serves as our primary diversity metric, while the VideoMAE~V2 counterpart and the $\ell_2$ metrics act as cross-checks. Where scale-free comparisons are needed, we report the teacher-normalized variant (teacher $=1.0$).
All metrics are computed on the same $P$ prompts and the same $N{=}5$ initial noise seeds for every method, inducing paired per-prompt measurements; method differences are assessed with a paired bootstrap over prompts ($10{,}000$ resamples).

\begin{table}[t]
\centering
\small
\setlength{\tabcolsep}{2.5pt} 
\renewcommand{\arraystretch}{1.2}

\caption{
VBench comparison on Wan2.1-T2V at the 1.3B and 14B scales.
All methods are evaluated at $480\times832$ resolution.
Bold values indicate the best result among the four-step students in each column for the corresponding model scale; the full-step teacher row and the NFE$=2$ row are reported as references and are excluded from bolding.
For the full-step teacher, $50\times2$ denotes 50 denoising steps with classifier-free guidance.
The NFE$=2$ row reports CrossDistill under a stricter two-step budget on the 1.3B model.
CrossDistill is the proposed trajectory-level scheduling policy; the rows below report its default instantiation $\mathrm{PCM}^{\mathrm{H}}+\mathrm{DMD}^{\mathrm{L}}$ (see note).
}

\label{tab:vbench_main}

\arrayrulecolor{tridentRed}

\begin{tabular}{l c c c c c c c c c c}
\toprule[1.25pt]

\multirow{2}{*}{\textbf{Method}} &
\multirow{2}{*}{\textbf{Params}} &
\multirow{2}{*}{\textbf{Resolution}} &
\multirow{2}{*}{\textbf{NFE} $\downarrow$} &
\multicolumn{3}{c}{\textbf{VBench}} &
\multicolumn{2}{c}{\textbf{V-JEPA 2}} &
\multicolumn{2}{c}{\textbf{VideoMAE V2}} \\
\cmidrule(lr){5-7} \cmidrule(lr){8-9} \cmidrule(lr){10-11}
 & & & & \textbf{Quality} $\uparrow$ & \textbf{Semantic} $\uparrow$ & \textbf{Total} $\uparrow$ & \textbf{Cos} $\uparrow$ & \textbf{L2} $\uparrow$ & \textbf{Cos} $\uparrow$ & \textbf{L2} $\uparrow$ \\

\midrule[0.65pt]

\rowcolor{tridentLight}
Wan2.1-T2V-1.3B & 1.3B & $480\times832$ & $50\times2$ & 85.09 & 75.73 & 83.22 & 0.124 & 27.09 & 0.0271 & 2.91 \\

PCM & 1.3B & $480\times832$ & 4 & 83.15 & 75.80 & 81.68 & \best{0.113} & \best{24.88} & 0.0196 & 2.58 \\
rCM & 1.3B & $480\times832$ & 4 & 84.87 & 75.54 & 83.00 & 0.105 & 24.18 & 0.0172 & 2.37 \\
AnyFlow & 1.3B & $480\times832$ & 4 & 85.16 & 76.42 & 83.41 & 0.082 & 19.92 & 0.0133 & 2.15 \\
DMD & 1.3B & $480\times832$ & 4 & 85.32 & 73.78 & 83.01 & 0.071 & 20.18 & 0.0144 & 2.18 \\

\rowcolor{tridentLight}
\textcolor{tridentRed}{\textbf{CrossDistill}\textsuperscript{$\dagger$}} & 1.3B & $480\times832$ & 4 & \best{85.41} & \best{77.21} & \best{83.77} & 0.106 & 24.70 & \best{0.0195} & \best{2.64} \\

\rowcolor{tridentLight}
\textcolor{tridentRed}{\textbf{CrossDistill}\textsuperscript{$\dagger$}} & 1.3B & $480\times832$ & 2 & 84.78 & 75.77 & 82.98 & 0.108 & 24.81 & 0.0192 & 2.51 \\

\arrayrulecolor{tridentOrange}
\specialrule{1.2pt}{3pt}{3pt}
\arrayrulecolor{tridentRed}

\rowcolor{tridentLight}
Wan2.1-T2V-14B & 14B & $480\times832$ & $50\times2$ & 85.97 & 76.68 & 84.11 & 0.125 & 27.15 & 0.0252 & 2.83 \\

PCM & 14B & $480\times832$ & 4 & 84.60 & 78.10 & 83.30 & \best{0.098} & \best{23.93} & 0.0152 & 2.54 \\
rCM & 14B & $480\times832$ & 4 & 85.62 & 77.24 & 83.94 & 0.082 & 22.09 & 0.0129 & 2.12 \\
AnyFlow & 14B & $480\times832$ & 4 & 85.77 & 77.58 & 84.13 & 0.078 & 21.00 & 0.0128 & 2.01 \\
DMD & 14B & $480\times832$ & 4 & \best{86.06} & 74.84 & 83.82 & 0.056 & 18.27 & 0.0084 & 1.61 \\

\rowcolor{tridentLight}
\textcolor{tridentRed}{\textbf{CrossDistill}\textsuperscript{$\dagger$}} & 14B & $480\times832$ & 4 & 85.96 & \best{78.44} & \best{84.46} & 0.095 & 23.88 & \best{0.0154} & \best{2.57} \\

\bottomrule[1.25pt]
\end{tabular}

\arrayrulecolor{black}

\vspace{2pt}
\begin{minipage}{0.98\linewidth}
\footnotesize
\textsuperscript{$\dagger$}CrossDistill denotes the scheduling policy with default instantiation $\mathrm{PCM}^{\mathrm{H}}+\mathrm{DMD}^{\mathrm{L}}$ (alternatives
in Table~\ref{tab:assign}). Diversity columns: raw cosine/$\ell_2$ distances in frozen
V-JEPA~2 and VideoMAE~V2 embeddings, $N{=}5$ seeds per prompt; higher is more diverse. Under the paired bootstrap, the Total-score improvement over the strongest four-step baseline is significant ($p<0.05$) at both scales, and the diversity improvements over DMD and AnyFlow are significant at $p<0.01$.

\end{minipage}
\end{table}

\subsection{Main Results}
\label{sec:main}

Table~\ref{tab:vbench_main} reports VBench results on Wan2.1-T2V, where CrossDistill corresponds to the default instantiation $\mathrm{PCM}^{\mathrm{H}}+\mathrm{DMD}^{\mathrm{L}}$.
 CrossDistill achieves the best semantic and total scores among four-step students at both scales; notably, at 1.3B its total score surpasses the full-step teacher. While monolithic DMD obtains the highest raw quality score, it suffers a significant semantic drop. We anchor diversity to the teacher (normalized $=1.0$) rather than maximizing it: CrossDistill recovers $\approx\!86\%$ of the teacher's seed-level diversity, versus $57\%$ for DMD and $66\%$ for AnyFlow at the 1.3B scale. Under the stricter NFE$=2$ budget, CrossDistill remains highly competitive in total scores with only a marginal quality degradation.

\begin{figure}[t]
    \centering

    \includegraphics[
        width=0.87\textwidth
    ]{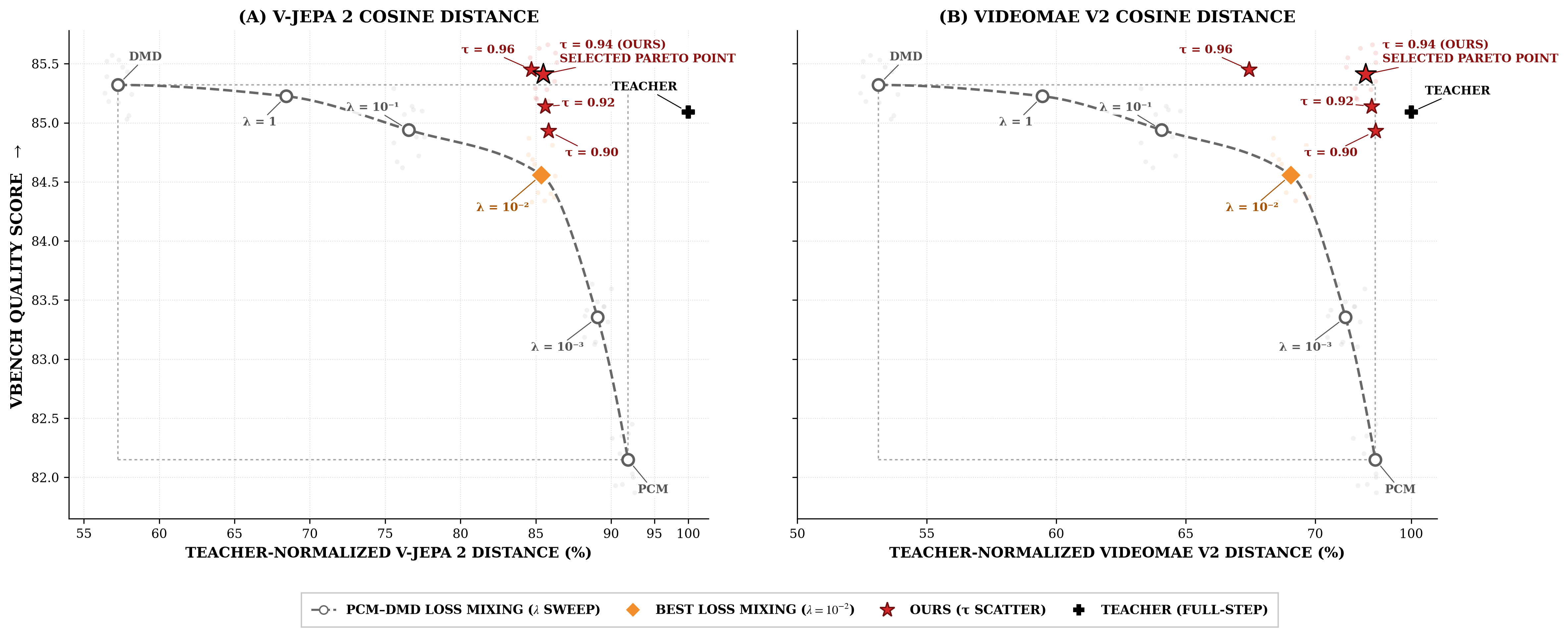}

    \vspace{-2pt}

    \captionsetup{
        font=footnotesize,
        skip=2pt
    }
\caption{
Quality--diversity comparison under four-step generation (NFE=$4$) on Wan2.1-T2V-1.3B.
Quality is the VBench Quality Score and Diversity is the teacher-normalized V-JEPA~2 cosine distance (the primary diversity metric; the corresponding VideoMAE~V2 diversity scores are reported in Tables~\ref{tab:vbench_main}--\ref{tab:assign}).
Gray circles represent PCM--DMD loss-level mixing with $\lambda_{\mathrm{PCM}}=1$ and
$\lambda_{\mathrm{DMD}}\in\{1,0.1,0.01,0.001\}$.
Red stars represent CrossDistill with $\tau^\star\in\{0.90,0.92,0.94,0.96\}$.
Faint points show prompt-level results, large markers report aggregate means, and the shaded region shows the range of prompt-level results.
Higher values are better on both axes.
}

    \label{fig:qd_pareto_mix}

\end{figure}

Fig.~\ref{fig:qd_pareto_mix} compares loss-level mixing and CrossDistill under the same four-step budget. Changing $\lambda_{\mathrm{DMD}}$ moves the loss-level baseline along a quality--diversity curve: larger $\lambda_{\mathrm{DMD}}$ improves quality but reduces diversity, while smaller $\lambda_{\mathrm{DMD}}$ has the opposite effect. In this sweep, CrossDistill extends the frontier toward the upper-right region, with $\tau^\star=0.94$ providing a favorable balance.

 \setlength{\columnsep}{10pt}
\setlength{\intextsep}{2pt}

\setlength{\columnsep}{10pt}
\setlength{\intextsep}{2pt}

\begin{figure}[h]
    \centering
    \includegraphics[width=\linewidth]{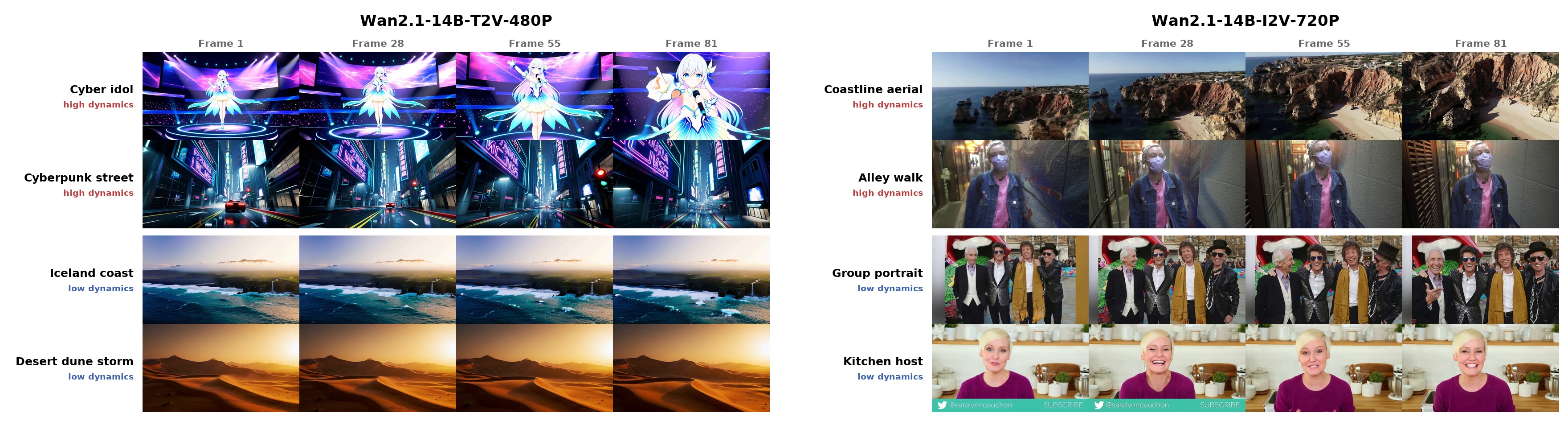}
\caption{Qualitative T2V and I2V results. We show Wan2.1-14B T2V-480P (left) and Wan2.1-14B I2V-720P (right). Each row corresponds to one prompt, with four frames uniformly sampled from an 81-frame sequence. The prompts include two high-motion cases and two near-static cases.}

    \label{fig:quality_demo_t2v_i2v}
\end{figure}

\begin{figure}[H]
    \centering
    \includegraphics[
        width=0.85\linewidth
    ]{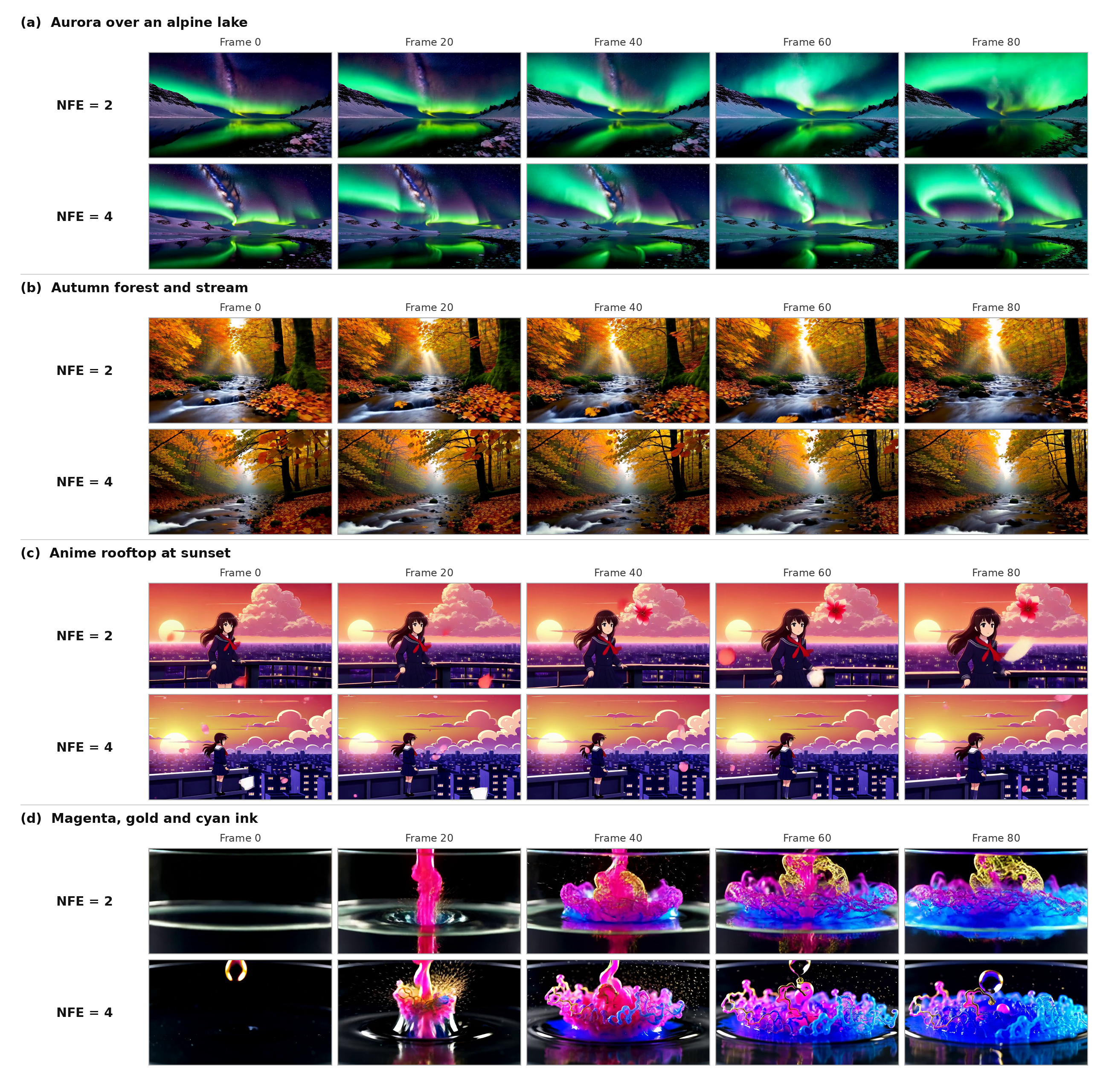}

    \caption{\textbf{Qualitative comparison of CrossDistill under the Wan2.1-1.3B 480p setting.}
    Each example shows four uniformly sampled frames from an 81-frame video, with the top and
    bottom rows corresponding to 2-NFE and 4-NFE generation, respectively.
    Across diverse scenes, the 2-NFE model preserves strong visual quality, semantic fidelity,
    and temporal consistency comparable to the 4-NFE counterpart, demonstrating an effective
    quality--efficiency trade-off.}
    \label{fig:CrossDistill_NFE2_NFE4_comparison}
\end{figure}

Qualitatively, Fig.~\ref{fig:diversity_comparison_wide} visualizes the quality--diversity trade-off across baselines: DMD and AnyFlow are sharp but map different seeds to nearly identical content, whereas rCM retains seed-level variation with degraded quality; CrossDistill improves the trade-off.
Fig.~\ref{fig:quality_demo_t2v_i2v} further shows 14B results on T2V (480P) and I2V (720P): CrossDistill preserves temporal coherence on high-motion prompts without visible drift or ghosting, and reduces the collapse of near-static prompts into frozen frames. This matches the stage assignment: high-noise TD preserves motion and semantic branching, while low-noise DM refines local details.
Appendix~\ref{sec:additional_results} provides further qualitative results, including NFE$=2$ vs.\ NFE$=4$ comparisons (Fig.~\ref{fig:CrossDistill_NFE2_NFE4_comparison}) showing that the crossover design remains effective and preserves semantic alignment under a tighter budget.

\subsection{Ablation Studies}
\label{sec:ablation}

\begin{wrapfigure}[18]{r}{0.45\textwidth}
    \vspace{0pt}
    \centering

    \includegraphics[
        width=\linewidth
    ]{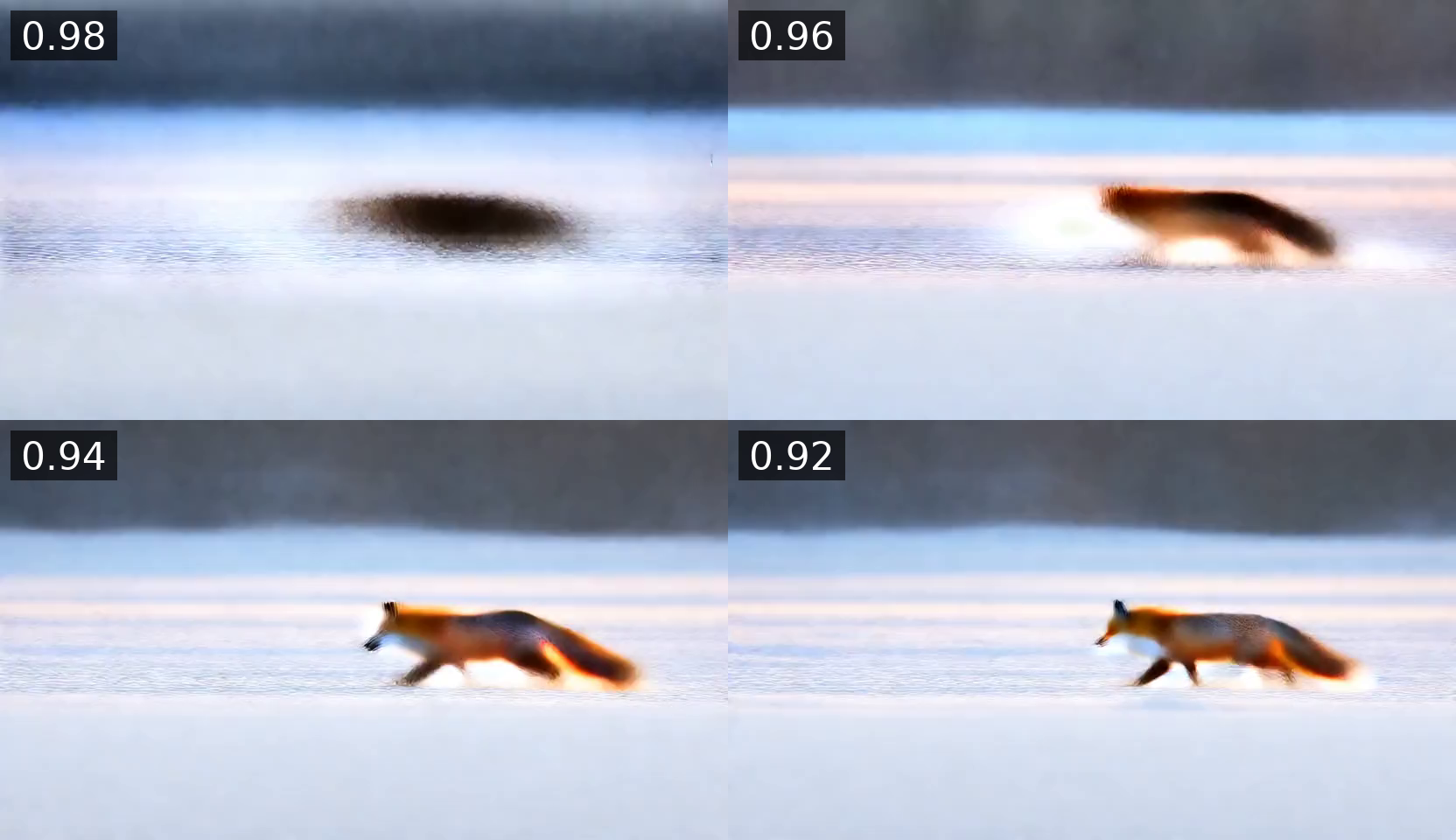}

    \vspace{-2pt}

    \captionsetup{
        font=footnotesize,
        skip=4pt
    }
     \caption{
    Effect of the crossover level on Wan2.1-T2V-1.3B.
    We visualize the intermediate states of the multi-step teacher denoised to different noise levels $\tau$.
    Although showing a single instance, this sample is representative of a consistent pattern observed across diverse prompts: around $\tau \approx 0.94$, the global mode reliably transitions from a noise-dominated state to a clear semantic structure. This observation identifies a generalizable boundary between high-noise mode selection and low-noise detail refinement.
    }

    \label{fig:crossover_grid_visual}

    \vspace{-0.3\baselineskip}
\end{wrapfigure}

\textbf{Crossover level $\tau^\star$.}
We investigate the sensitivity of the crossover level $\tau^\star$ on Wan2.1-1.3B. To determine a robust boundary between mode selection and detail refinement, we first visualize the intermediate denoising states across a diverse set of prompts and random seeds. As consistently shown in Fig.~\ref{fig:crossover_grid_visual}, the global semantic mode and coarse motion dynamics largely emerge from the noise around $\tau \approx 0.94$, regardless of the specific text condition. This cross-prompt consistency suggests that $\tau \approx 0.94$ serves as a generalizable phase-transition boundary: before this point, trajectory distillation is critical for preserving input-dependent variation; after this point, the committed branch primarily requires local manifold refinement. Guided by this qualitative observation, we conduct a grid search over $\tau^\star$ to identify the optimal operating point. The quantitative results in Fig.~\ref{fig:qd_pareto_mix} confirm that $\tau^\star=0.94$ yields the best quality--diversity balance, extending the Pareto frontier furthest toward the upper-right region. This validates that our default crossover level is not prompt-specific, but a stable scheduling prior for few-step distillation of Wan2.1.

\setlength{\columnsep}{10pt}
\setlength{\intextsep}{2pt}

\begin{figure}[H]
    \centering
    \includegraphics[
        width=0.88\linewidth
    ]{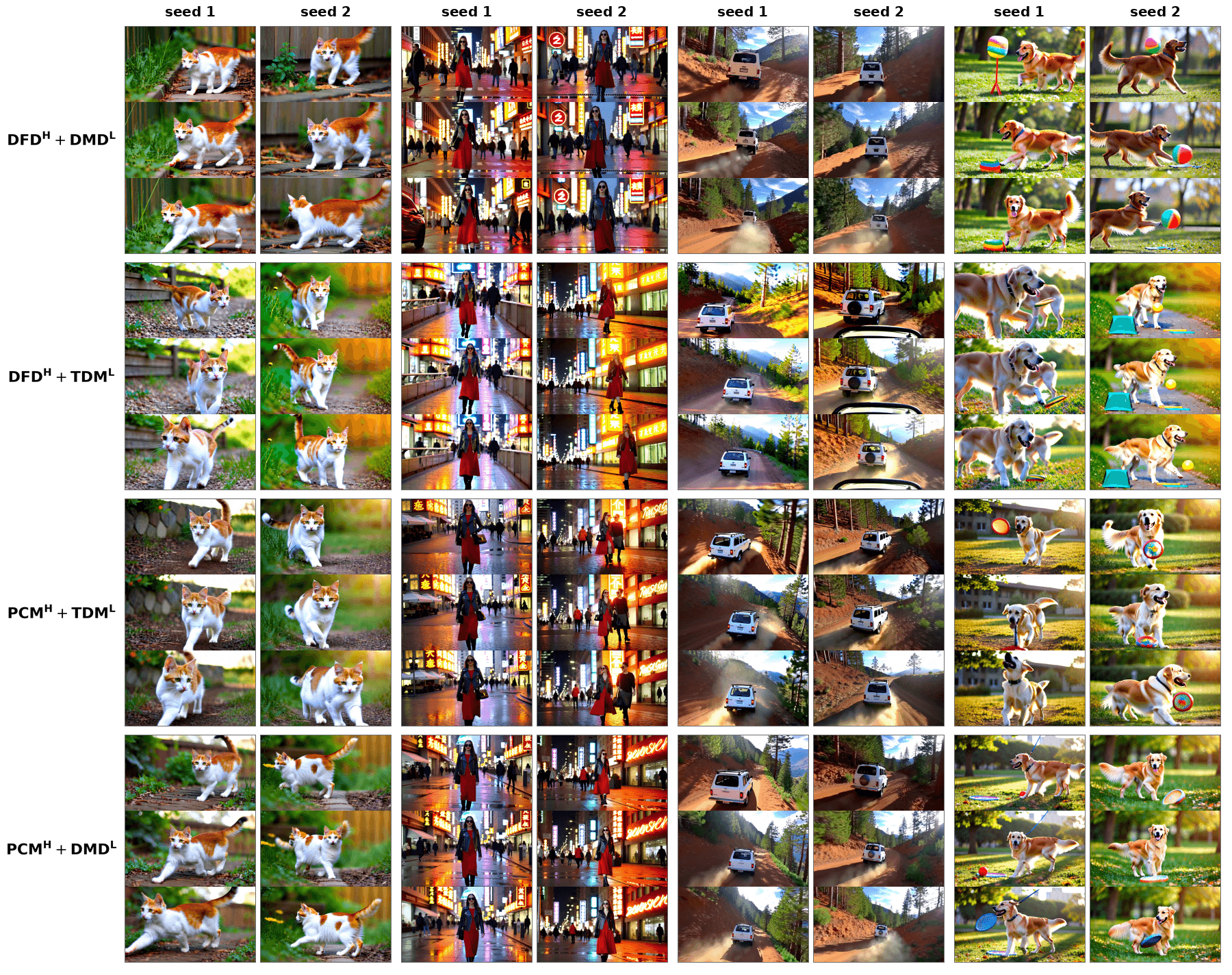}

    \caption{
    Qualitative comparison across instantiations of the CrossDistill policy
    under the Wan2.1-1.3B 4-step budget.
    Rows correspond to the four configurations in Table~\ref{tab:assign}:
    $\mathrm{DFD}^{\mathrm{H}}+\mathrm{TDM}^{\mathrm{L}}$,
    $\mathrm{PCM}^{\mathrm{H}}+\mathrm{TDM}^{\mathrm{L}}$,
    $\mathrm{DFD}^{\mathrm{H}}+\mathrm{DMD}^{\mathrm{L}}$, and
    $\mathrm{PCM}^{\mathrm{H}}+\mathrm{DMD}^{\mathrm{L}}$.
    Columns show generated samples across different random seeds.
    }
    \label{fig:comparison_4methods_3frames}
\end{figure}

\textbf{Stage assignment: instantiations of the same policy.}
Table~\ref{tab:assign} and Fig.~\ref{fig:comparison_4methods_3frames} instantiate the CrossDistill scheduling policy with different objective pairs while keeping the crossover partition, relay interface, and training protocol fixed. We combine two high-noise TD objectives, $\mathrm{DFD}^{\mathrm{H}}$ and $\mathrm{PCM}^{\mathrm{H}}$, with two low-noise DM objectives, $\mathrm{TDM}^{\mathrm{L}}$ and $\mathrm{DMD}^{\mathrm{L}}$. As shown in Fig.~\ref{fig:comparison_4methods_3frames} and Table~\ref{tab:assign}, all four tested instantiations achieve a favorable quality--diversity trade-off under our protocol. This suggests that the benefit of CrossDistill comes from the noise-level schedule rather than a single loss pair, and is robust to the tested TD/DM instantiations. The main-table model, $\mathrm{PCM}^{\mathrm{H}}+\mathrm{DMD}^{\mathrm{L}}$, is one default instantiation of this policy.

\begin{table}[t]
\centering
\small
\setlength{\tabcolsep}{4pt}
\renewcommand{\arraystretch}{1.2}

\caption{
Instantiations of the CrossDistill scheduling policy with identical crossover partition and training protocol.
The superscripts $\mathrm{H}$ and $\mathrm{L}$ denote the objective used on the high-noise and low-noise intervals, respectively.
All methods are evaluated at $480\times832$ resolution on the 1.3B model.
Bold values indicate the best result among the configurations.
The $\mathrm{PCM}^{\mathrm{H}}+\mathrm{DMD}^{\mathrm{L}}$ row coincides with the CrossDistill row (1.3B, NFE$=4$) of Table~\ref{tab:vbench_main}.
}

\label{tab:assign}

\arrayrulecolor{tridentRed}

\begin{tabular}{l c c c c c c c}
\toprule[1.25pt]

\multirow{2}{*}{\textbf{Configuration}} &
\multicolumn{3}{c}{\textbf{VBench}} &
\multicolumn{2}{c}{\textbf{V-JEPA 2}} &
\multicolumn{2}{c}{\textbf{VideoMAE V2}} \\
\cmidrule(lr){2-4} \cmidrule(lr){5-6} \cmidrule(lr){7-8}
 & \textbf{Quality} $\uparrow$ & \textbf{Semantic} $\uparrow$ & \textbf{Total} $\uparrow$ & \textbf{Cos} $\uparrow$ & \textbf{L2} $\uparrow$ & \textbf{Cos} $\uparrow$ & \textbf{L2} $\uparrow$ \\

\midrule[0.65pt]

CrossDistill ($\mathrm{DFD}^{\mathrm{H}}+\mathrm{TDM}^{\mathrm{L}}$) & 85.06 & 76.17 & 83.28 & 0.099 & 23.50 & 0.0169 & 2.39 \\

CrossDistill ($\mathrm{PCM}^{\mathrm{H}}+\mathrm{TDM}^{\mathrm{L}}$) & 85.09 & 76.85 & 83.44 & 0.104 & 24.49 & 0.0181 & 2.49 \\

CrossDistill ($\mathrm{DFD}^{\mathrm{H}}+\mathrm{DMD}^{\mathrm{L}}$) & 85.20 & 77.04 & 83.57 & 0.098 & 23.46 & 0.0169 & 2.37 \\

\rowcolor{tridentLight}
\textcolor{tridentRed}{\textbf{CrossDistill ($\mathrm{PCM}^{\mathrm{H}}+\mathrm{DMD}^{\mathrm{L}}$)}} & \best{85.41} & \best{77.21} & \best{83.77} & \best{0.106} & \best{24.70} & \best{0.0195} & \best{2.64} \\

\bottomrule[1.25pt]
\end{tabular}

\arrayrulecolor{black}

\vspace{2pt}
\begin{minipage}{0.98\linewidth}
\footnotesize
$\mathrm{DFD}^{\mathrm{H}}$ is the one-step high-noise $\ell_2$ regression to the teacher-induced crossover state (Eq.~\eqref{eq:dfd_high}).
$\mathrm{PCM}^{\mathrm{H}}$ is the high-noise consistency relay (Eq.~\eqref{eq:pcm}).
$\mathrm{TDM}^{\mathrm{L}}$ and $\mathrm{DMD}^{\mathrm{L}}$ are low-noise distribution-matching objectives; $\mathrm{DMD}^{\mathrm{L}}$ uses Eq.~\eqref{eq:dmd}, and $\mathrm{TDM}^{\mathrm{L}}$ uses the TDM objective~\cite{tdm} on the same low-noise interval.
The default model in Table~\ref{tab:vbench_main} is $\mathrm{PCM}^{\mathrm{H}}+\mathrm{DMD}^{\mathrm{L}}$.
\end{minipage}
\end{table}

\textbf{Reversed stage assignment.}
The preceding stage-assignment study changes the concrete TD/DM losses while preserving the proposed ordering. We additionally test whether the ordering itself matters. To this end, we implement a reversed schedule that keeps the same crossover point $\tau^\star$ and NFE budget, but swaps the noise supports of the two objectives: the DM objective is sampled only on the high-noise interval $\mathcal{H}$, while the TD objective is sampled only on the low-noise interval $\mathcal{L}$. This control helps isolate the coarse-to-fine responsibility: if high-noise TD is responsible for preserving mode branching and low-noise DM is responsible for local fidelity, reversing the two should reduce diversity and should not improve quality.

\begin{table}[htbp]
\centering
\small
\setlength{\tabcolsep}{4pt}
\renewcommand{\arraystretch}{1.2}

\caption{
Reversed stage assignment on Wan2.1-T2V-1.3B.
All methods use NFE$=4$ and $\tau^\star=0.94$.
The reversed variant keeps the same crossover partition and training protocol but swaps the noise supports of the TD and DM objectives.
}

\label{tab:reversed_assignment}

\arrayrulecolor{tridentRed}

\begin{tabular}{l c c c c c c c}
\toprule[1.25pt]

\multirow{2}{*}{\textbf{Configuration}} &
\multicolumn{3}{c}{\textbf{VBench}} &
\multicolumn{2}{c}{\textbf{V-JEPA 2}} &
\multicolumn{2}{c}{\textbf{VideoMAE V2}} \\
\cmidrule(lr){2-4} \cmidrule(lr){5-6} \cmidrule(lr){7-8}
 & \textbf{Quality} $\uparrow$ & \textbf{Semantic} $\uparrow$ & \textbf{Total} $\uparrow$ & \textbf{Cos} $\uparrow$ & \textbf{L2} $\uparrow$ & \textbf{Cos} $\uparrow$ & \textbf{L2} $\uparrow$ \\

\midrule[0.65pt]

CrossDistill ($\mathrm{PCM}^{\mathrm{H}}+\mathrm{DMD}^{\mathrm{L}}$) & 85.41 & 77.21 & 83.77 & 0.106 & 24.70 & 0.0195 & 2.64 \\

\rowcolor{tridentLight}
Reversed ($\mathrm{DMD}^{\mathrm{H}}+\mathrm{PCM}^{\mathrm{L}}$) & 82.42 & 75.93 & 81.12 & 0.082 & 20.90 & 0.0152 & 2.13 \\

\bottomrule[1.25pt]
\end{tabular}

\arrayrulecolor{black}

\vspace{2pt}
\begin{minipage}{0.98\linewidth}
\footnotesize
The first row is the same as the 1.3B, NFE$=4$ result in Table~\ref{tab:vbench_main}. The reversed variant applies the DM objective on $\mathcal{H}$ and the TD objective on $\mathcal{L}$; all other settings, including base learning rates, NFE budget, and critic protocol, are kept the same.
\end{minipage}
\end{table}

\begin{figure}[h]
    \centering
    \includegraphics[width=\linewidth]{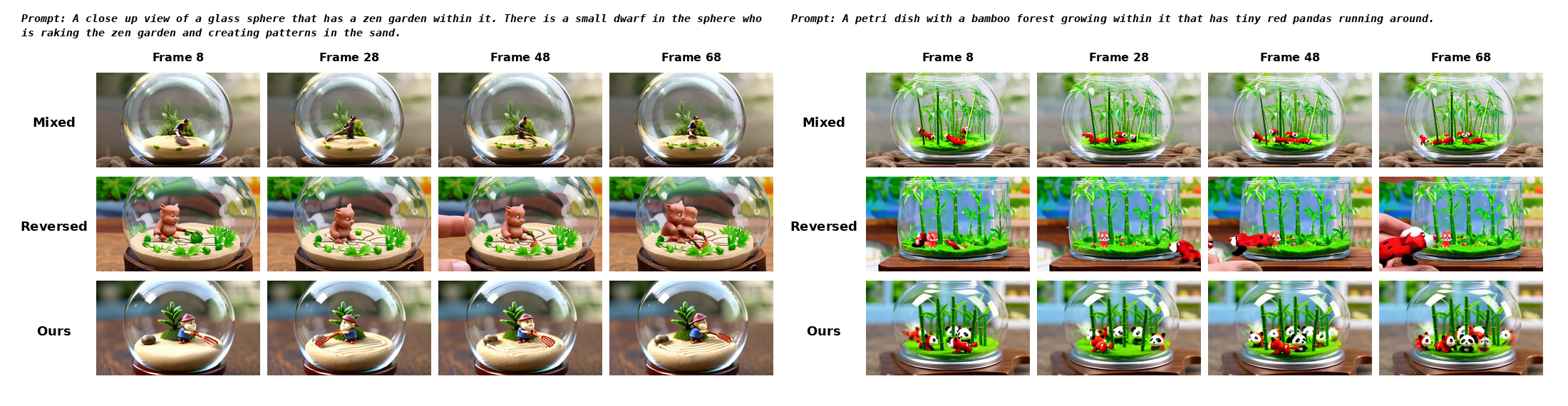}
        \caption{\textbf{Qualitative comparison of mixed, reversed, and proposed schedules.}
    We compare the tuned PCM--DMD loss-level mixture (Mixed), the reversed schedule
    $\mathrm{DMD}^{\mathrm{H}}+\mathrm{PCM}^{\mathrm{L}}$, and our default schedule
    $\mathrm{PCM}^{\mathrm{H}}+\mathrm{DMD}^{\mathrm{L}}$ under the same prompts,
    random seeds, and four-step budget. In these examples, the proposed schedule better preserves
    the intended prompt semantics, while both Mixed and Reversed show noticeable semantic
    deviations. This visualization provides qualitative support for the
    coarse-to-fine assignment of TD to high-noise mode selection and DM to low-noise
    detail refinement.}
    \label{fig:ours_mixed_reversed}
\end{figure}

Table~\ref{tab:reversed_assignment} reports the quantitative results for this control: the reversed variant does not recover the quality--diversity trade-off of CrossDistill in our setting. Its diversity is markedly lower, consistent with the hypothesis that high-noise DM may prematurely collapse input-dependent variation. Furthermore, its quality does not improve, suggesting that low-noise TD does not substitute for low-noise distribution matching in recovering local manifold fidelity under our protocol. The qualitative comparison in Fig.~\ref{fig:ours_mixed_reversed} provides a visualization of this failure mechanism. Under identical prompts and initial noises, both the reversed schedule and the tuned loss-level mixture deviate noticeably from the intended prompt semantics in these examples, whereas CrossDistill better preserves the global semantic structure. This suggests a possible causal chain: once high-noise mode selection is corrupted---whether by reversing the stage order or by blending the two objectives at every noise level via loss-level mixing---the discarded semantic alternatives and stochastic diversity may be difficult to recover at low noise. Overall, these results show that the proposed noise-level ordering, not merely the TD--DM combination, drives the improvement.


\section{Conclusion}
\label{sec:conclusion}

High-noise decisions tend to shape mode structure and diversity, while low-noise decisions tend to refine fidelity and local realism, so distillation objectives can be scheduled according to the role of each noise regime. CrossDistill formalizes this idea as a trajectory-level scheduling policy with a single crossover between trajectory-based distillation and distribution matching. The default $\mathrm{PCM}^{\mathrm{H}}+\mathrm{DMD}^{\mathrm{L}}$ model is one instantiation of this policy; the stage-assignment and reversed-schedule results suggest that the noise-level ordering, not merely the TD--DM combination, is the key factor. More broadly, this work treats the denoising trajectory as a structured design space in which complementary objectives can be composed at the trajectory level.

\section{Acknowledgments}
This work was supported by Alibaba Group through Alibaba Research Intern Program.

\clearpage
\beginappendix

\section{Related Work}
\label{sec:related}

\textbf{Few-step diffusion distillation.}
Diffusion and flow matching models achieve strong generation quality but typically require many sampling steps~\cite{ho2020denoising,song2020denoising,lipman2022flow,liu2022flow,esser2024scaling}. Few-step distillation reduces this cost by training a student model to approximate the teacher's generative behavior with a small number of function evaluations. Existing objectives are often designed around either trajectory preservation or distribution alignment. Our work is not focused on proposing yet another monolithic distillation loss; instead, it makes the noise axis an explicit design variable and studies where along the sampling trajectory different distillation objectives should act under a fixed few-step budget.

\textbf{Trajectory-based distillation.}
Trajectory-based distillation trains a student to preserve the teacher's transport trajectories over fewer steps. Progressive distillation mimics a teacher over progressively shorter sampling intervals~\cite{salimans2022progressive}. Consistency models enforce that states on the same probability-flow ODE trajectory map to a common output~\cite{song2023consistency}, with subsequent work improving training stability and sample quality~\cite{song2024improved,scm,luo2023lcm}. Consistency trajectory models extend this idea to mappings between arbitrary noise levels~\cite{kim2024consistency}, and phased consistency models partition the trajectory into phases for few-step sampling~\cite{pcm}. These objectives are useful for preserving mode coverage and seed-level variation, but with very few steps they may not fully recover local data-manifold fidelity, often leading to blurred or over-smoothed samples. This limitation suggests that trajectory preservation alone may not be equally valuable at every noise level.

\textbf{Distribution matching and adversarial distillation.}
Another line of work aligns the student's marginal distribution directly with the teacher's. DMD uses a score-difference gradient to minimize a distributional divergence~\cite{yin2024one}, and DMD2 stabilizes distribution matching with a two-time-scale update rule and adversarial supervision~\cite{yin2024improved}. TDM incorporates trajectory awareness into distribution matching~\cite{tdm}. SiD derives a score-identity objective for one-step generation~\cite{zhou2024score}, and Diff-Instruct transfers knowledge by minimizing an integral KL-type objective~\cite{luo2023diff}. 
Recent DMD-style variants further improve stability and scaling~\cite{senseflow,decoupleddmd,cdmd}. These methods often produce sharper samples, but because they act primarily on marginals rather than individual trajectories, they can become mode-seeking and reduce stochastic diversity. In video generation, this issue is especially visible as reduced seed-level variation and weakened temporal dynamics. The key question, therefore, is not whether distribution matching is useful, but \emph{in which noise regimes it should be allowed to act}.

\textbf{Hybrid objectives and staged distillation.}
Recent distillation recipes increasingly involve multiple objectives. rCM couples consistency training with distribution-matching-style supervision~\cite{rcm}, while other methods introduce auxiliary adversarial or regularization terms to stabilize few-step training~\cite{yin2024improved,sauer2024adversarial,dmdr}. Other approaches separate objectives in training time: AnyFlow~\cite{gu2026anyflow} applies an off-policy flow-map stage followed by an on-policy DMD stage, and \textit{From Structure to Detail}~\cite{Cheng2025From} post-trains a trajectory-distilled student with distribution matching. These methods demonstrate that trajectory- and distribution-based objectives are complementary. However, in most such recipes the objectives are either mixed over broad noise ranges or separated only by training stage, leaving the effective noise regime of each objective implicit. CrossDistill differs by making the noise axis an explicit scheduling variable: trajectory-based distillation is restricted to the high-noise mode-commitment interval, while distribution matching is restricted to the low-noise detail-refinement interval. This trajectory-level partition is orthogonal to training-time sequencing and can be instantiated with different plug-in objectives.

\textbf{Coarse-to-fine denoising and noise-dependent generation.}
Several studies observe that diffusion sampling proceeds from coarse to fine: high-noise steps are associated with global layout, semantics, and mode selection, whereas low-noise steps refine textures and local details~\cite{meng2021sdedit,choi2021ilvr,hertz2022prompt}. Noise-scheduling analyses further show that different noise levels affect different aspects of generation~\cite{ho2020denoising,chen2023noisescheduling}. Unlike prior work that uses coarse-to-fine behavior mainly for editing, conditioning, or base-model scheduling, we use it to derive an objective-assignment rule for few-step distillation. We further test this rule with controlled ablations, including reversed schedules and multiple objective instantiations, rather than treating it as a heuristic curriculum.

\section{Conceptual Discussion: A Two-Price View of Noise-Level Scheduling}

\label{app:insight}

This appendix provides a simplified conceptual view of the quality--diversity trade-off. It is not a formal theory, proof, or model of optimization dynamics. We use it to clarify the intuition behind the proposed noise-level schedule; all empirical claims in the paper are supported by the experiments in the main text.

\paragraph{A simplified bookkeeping view.}
For a schedule $\alpha:[0,1]\to[0,1]$ giving the TD weight at noise level $t$, write two
reduced-form prices,
$w_D(t)$ = diversity cost of using DM instead of TD at $t$, and
$w_Q(t)$ = quality cost of using TD instead of DM at $t$,
and let the total bills add over levels:
\begin{equation}
  D(\alpha)=\int_0^1 \bigl(1-\alpha(t)\bigr) w_D(t)\,dt,
  \qquad
  Q(\alpha)=\int_0^1 \alpha(t)\, w_Q(t)\,dt .
  \label{eq:bills}
\end{equation}
Additivity ignores same-level interactions between the two losses. This is related to what
Fig.~\ref{fig:qd_pareto_mix} examines: in our sweep, tuned uniform blends approximately trace the chord between the two
monolithic students and do not fall below it, so we do not observe strong same-level synergy within this limited sweep.

\paragraph{Coarse-to-fine intuition.}
The intuition behind the rule is the ordering of the exchange rate $R(t):=w_D(t)/w_Q(t)$: diversity is
expensive to lose at high noise and cheap at low noise, while quality is cheap to lose at
high noise and expensive at low noise. This ordering is consistent with prior observations: high-noise steps
fix global layout, semantics, and mode identity while low-noise steps refine texture and
local statistics~\cite{ho2020denoising,meng2021sdedit,choi2021ilvr,hertz2022prompt,chen2023noisescheduling}.
Two anchors in our own experiments are consistent with it: at $t\approx0.94$ a purely
DM-distilled student already maps distinct seeds to nearly identical states
(Fig.~\ref{fig:prompt0_comparison}), suggesting that $w_D$ is large at high noise; and low-noise TD
neither restores manifold fidelity nor improves quality over low-noise DM
(Fig.~\ref{fig:trajectory_alignment}, Table~\ref{tab:reversed_assignment}), suggesting that $w_Q$ is
large at low noise. We do not claim $R$ is exactly monotone in practice; the ordering
defines the regime in which the rule is intended to operate.

\paragraph{A two-level illustration.}
The mechanism is visible with one high-noise level $H$ and one low-noise level $L$ and
prices $w_D(H)=w_Q(L)=1$, $w_D(L)=w_Q(H)=0.1$:
\begin{center}
\small
\begin{tabular}{@{}llccc@{}}
\toprule
schedule & prices paid & $D(\alpha)$ & $Q(\alpha)$ & $u+v$ \\
\midrule
TD everywhere & $w_Q(H)+w_Q(L)$ & $0$ & $1.1$ & $1$ \\
DM everywhere & $w_D(H)+w_D(L)$ & $1.1$ & $0$ & $1$ \\
uniform blend & half of each price & $0.55$ & $0.55$ & $1$ \\
\textbf{crossover} (TD on $H$, DM on $L$) & the two cheap prices & $\mathbf{0.1}$ & $\mathbf{0.1}$ & $\mathbf{0.18}$ \\
reversed (DM on $H$, TD on $L$) & the two expensive prices & $1.0$ & $1.0$ & $1.82$ \\
\bottomrule
\end{tabular}
\end{center}
In this idealized example, the crossover pays the two cheap prices, while the reversed order pays the two expensive ones
and can be worse than uniform blends; uniform blends sweep the chord between the
monolithic endpoints. The continuum ledger gives an analogous picture, with the chord gap
equal to the area between the two normalized price curves.

\paragraph{Informal interpretation.}
Within this simplified bookkeeping view: (i) at a fixed noise level, blending corresponds to a convex combination of the two idealized costs; (ii) uniform blends trace a one-parameter trade-off between the two monolithic endpoints; and (iii) if the relative cost favors TD at high noise and DM at low noise, a single crossover becomes a simple and natural schedule. These statements are intended only as intuition and should not be read as guarantees for real training dynamics.

\paragraph{Scope and limitations.}
The ledger is bookkeeping, not dynamics: it prices which objective drives each level
regardless of training order, and says nothing about gradient interference or the
stop-gradient choice at $\tau^\star$. Three observations would weaken this picture, and two
are already tested: a tuned uniform blend below the chord (not observed,
Fig.~\ref{fig:qd_pareto_mix}) would challenge the additivity approximation; a reversed schedule on par with
ours (not observed, Table~\ref{tab:reversed_assignment}) would challenge the proposed ordering; a
strongly non-monotone measured $R$ would motivate considering multi-switch schedules---estimating $R$
level-wise (perturb each level's objective, read off diversity/quality deltas) is a
straightforward direction for future work.

\newpage

\section{Additional Results}
\label{sec:additional_results}

\subsection{Toy case}

\paragraph{Toy model setup.} We consider a 2D spiral density-estimation task. The gray background in the figure shows the target data distribution. Both the teacher and student are the same 4-layer MLP velocity networks with sinusoidal time embedding. The teacher is trained from scratch with flow matching to learn the target distribution until convergence, and serves as the ground-truth ODE solver. Each student is then distilled from the teacher for 10K iterations with batch size 1024.
\paragraph{Baselines.} Trajectory-based distillation (TD) trains the student with the PCM loss on 4-step trajectories. Distribution matching (DM) trains the student with the DMD distribution-matching loss on 4-step trajectories. The loss-level mixture (Mix) combines the PCM and DMD losses at every timestep. Our trajectory-level hybrid (CrossDistill) applies PCM in high-noise regions and DMD in low-noise regions.

\paragraph{Observations.} Colored curves denote 4-step denoising trajectories from 12 random seeds, and black dots mark the final generated samples. The leftmost panel shows the teacher ODE solver for reference.
PCM preserves the teacher trajectory well, but the predicted data points are less accurate. DMD places final samples on the target manifold, yet without trajectory consistency the points collapse into specific regions and lose diversity. The loss-level mixture roughly follows the teacher trajectory, but some generated points fall outside the manifold. In contrast, in this toy setting our trajectory-level hybrid better preserves both the teacher sampling trajectory and the target distribution modes, with most final samples lying closer to the target manifold.

\subsection{Additional Qualitative Comparisons}
\label{sec:additional_qualitative_comparisons}

\vspace*{\fill}

\vspace*{\fill}

\clearpage

\vspace*{\fill}

\vspace*{\fill}

\clearpage

\newpage
\section{Training and Model Hyperparameters}
\label{app:hyperparameters}

We summarize the model and training configurations used in our experiments
in Table~\ref{tab:training_hyperparameters}. Unless otherwise specified,
the same CrossDistill configuration is used for the 1.3B and 14B models.

\begin{tikzpicture}[remember picture,overlay]
\node[anchor=center] at (current page.center) {%
\begin{minipage}{0.78\textwidth}
\centering
\small
\setlength{\tabcolsep}{8pt}
\renewcommand{\arraystretch}{1.15}

\captionof{table}{Training and model hyperparameters for CrossDistill.}
\label{tab:training_hyperparameters}

\arrayrulecolor{tridentRed}

\begin{tabular}{l c c}
\toprule[1.25pt]

\textbf{Hyperparameter} &
\textbf{Wan2.1-T2V-1.3B} &
\textbf{Wan2.1-T2V-14B} \\

\midrule[0.65pt]

Base model & Wan2.1-T2V-1.3B & Wan2.1-T2V-14B \\
Model parameters & 1.3B & 14B \\
Resolution & $480\times832$ & $480\times832$ \\
Student NFE & 4 & 4 \\
Number of frames & 81 & 81 \\

\arrayrulecolor{tridentOrange}
\specialrule{0.8pt}{3pt}{3pt}
\arrayrulecolor{tridentRed}

Crossover point $\tau^\star$ & 0.94 & 0.94 \\
High-noise interval & $[0.94,1]$ & $[0.94,1]$ \\
Low-noise interval & $[0,0.94]$ & $[0,0.94]$ \\
High-noise objective & PCM & PCM \\
Low-noise objective & DMD & DMD \\
High-noise student steps & 1 & 1 \\
Low-noise student steps & 3 & 3 \\
DMD gradient through PCM relay & No & No \\
PCM/DMD update-strength ratio & $10:1$ & $10:1$ \\

\arrayrulecolor{tridentOrange}
\specialrule{0.8pt}{3pt}{3pt}
\arrayrulecolor{tridentRed}

Optimizer & AdamW & AdamW \\
DMD base learning rate & $2\times10^{-6}$ & $2\times10^{-6}$ \\
PCM base learning rate & $2\times10^{-6}$ & $2\times10^{-6}$ \\
Weight decay & 0.01 & 0.01 \\
Global batch size & 16 & 16 (480P) / 8 (720P) \\
Training iterations & 10,000 & 10,000 \\
Gradient clipping & Disabled & Disabled \\
Training precision & BF16 & BF16 \\

\arrayrulecolor{tridentOrange}
\specialrule{0.8pt}{3pt}{3pt}
\arrayrulecolor{tridentRed}

Critic learning rate & $4\times10^{-7}$ & $4\times10^{-7}$ \\
Critic update ratio & 4:1 & 9:1 \\

\bottomrule[1.25pt]
\end{tabular}

\arrayrulecolor{black}
\end{minipage}
};
\end{tikzpicture}

\vfill
\newpage

\newpage
\section{Algorithm}
\label{sec:Algorithm}

This appendix gives the full training procedure for the default instantiation of Sec.~\ref{sec:construction}; gradient flow, stop-gradient at the crossover, and step allocation follow Sec.~\ref{sec:construction} and are not repeated here. The student, teacher, and critic networks are parameterized as velocity predictors: for a velocity network $F$, the clean prediction at time $t$ is $\hat{x}_0=x_t-tF(x_t,t,c)$. The critic is trained with the flow-matching objective of Eq.~\eqref{eq:fake_fm}, denoted $\mathcal{L}_{\mathrm{FM}}(F;\hat{x}_0,c)$, and $p_D=\mathcal{U}([0,1])$, so $p_D^{\mathrm{late}}$ in Algorithm~\ref{alg:pcm_dmd} is the uniform distribution on $\mathcal{L}$ used in Eq.~\eqref{eq:dmd}. The PCM update uses a $10\times$ update-strength multiplier relative to the DMD update. The complete pseudocode is as follows:

\begin{algorithm}[t]
\caption{CrossDistill Training: PCM--DMD Instantiation}
\label{alg:pcm_dmd}
\begin{algorithmic}[1]
\Require
Frozen teacher velocity network $F_{\phi}$;
student generator $G_{\theta}$;
critic/fake velocity network $F_{\psi}$;
few-step schedule
$\mathcal{S}=\{\tau_0,\tau_1,\ldots,\tau_K\}$,
where $\tau_0=1$, $\tau_1=\tau^\star$, and $\tau_K=0$;
student update frequency $f$;
loss weights $\lambda_{\mathrm{DMD}}$ and $\lambda_{\mathrm{PCM}}$.
\Ensure Trained few-step generator $G_{\theta}$.

\State Define the PCM interval
$\mathcal{I}_{\mathrm{PCM}}=[\tau_1,\tau_0]$

\State Define the later interval
$\mathcal{I}_{\mathrm{late}}=[\tau_K,\tau_1]$ and
$p_D^{\mathrm{late}}(t)=p_D(t\mid t\in\mathcal{I}_{\mathrm{late}})$,
where $p_D$ is the DMD noise sampling distribution

\State Initialize $\theta$ from the pretrained teacher
\State $\psi\gets\phi$
\Comment{Initialize critic/fake model from teacher}

\For{$i=0,\ldots,N-1$}

    \If{$i\bmod f\neq0$}
        \Statex \textbf{// Update critic/fake model}

        \State Sample condition $c$ and $z\sim\mathcal{N}(0,I)$

        \State $\hat{x}_0\gets
        \mathrm{sg}\left[
        \mathrm{Rollout}
        \left(
        G_{\theta},z,c,
        \tau_0\rightarrow\cdots\rightarrow\tau_K
        \right)
        \right]$

        \State $\mathcal{L}_{\mathrm{fake}}\gets
        \mathcal{L}_{\mathrm{FM}}
        \left(F_{\psi};\hat{x}_0,c\right)$
        \Comment{Eq.~\eqref{eq:fake_fm}}

        \State $\psi\gets
        \psi-\eta_{\psi}
        \nabla_{\psi}\mathcal{L}_{\mathrm{fake}}$

    \Else
        \Statex \textbf{// Update student generator}

        \State Sample condition $c$ and $z\sim\mathcal{N}(0,I)$

        \State Sample rollout length
        $m\sim\mathrm{Uniform}\{2,\ldots,K\}$

        \State $\hat{x}_0\gets
        \mathrm{Rollout}
        \left(
        G_{\theta},z,c,
        \tau_0\rightarrow\cdots
        \rightarrow\tau_{m-1}\rightarrow\tau_K
        \right)$
         \Comment{DMD supervises the final prediction; stop gradient at the high-noise relay}

        \Statex \textbf{// DMD on the low-noise interval}

        \State Sample
        $t_D\sim p_D^{\mathrm{late}}(t)$ and
        $\epsilon_D\sim\mathcal{N}(0,I)$

        \State $x_{t_D}\gets
        (1-t_D)\hat{x}_0+t_D\epsilon_D$

        \State $v_{\psi}\gets
        F_{\psi}(x_{t_D},t_D,c)$

        \State $v_{\phi}\gets
        F_{\phi}(x_{t_D},t_D,c)$
        \Comment{Teacher prediction with CFG}

        \State $\hat{x}_0^{\psi}\gets
        x_{t_D}-t_D v_{\psi}$

        \State $\hat{x}_0^{\phi}\gets
        x_{t_D}-t_D v_{\phi}$

        \State $Z\gets
        \mathrm{mean}\big(
        |\hat{x}_0-\hat{x}_0^{\phi}|
        \big)$
        \Comment{Stop gradient}

        \State $g_{\mathrm{DMD}}\gets
        \dfrac{
        \hat{x}_0^{\psi}-\hat{x}_0^{\phi}
        }{
        Z
        }$

        \State $\mathcal{L}_{\mathrm{DMD}}\gets
        \left\|
        \hat{x}_0-
        \mathrm{sg}
        \left(
        \hat{x}_0-g_{\mathrm{DMD}}
        \right)
        \right\|_2^2$

        \Statex \textbf{// PCM supervision on the high-noise interval}

        \State Sample real latent $x_0$,
        $\epsilon\sim\mathcal{N}(0,I)$, and
        $t_a\in\mathcal{I}_{\mathrm{PCM}}$

        \State $x_{t_a}\gets
        (1-t_a)x_0+t_a\epsilon$

        \State Sample $t_b<t_a$ within
        $\mathcal{I}_{\mathrm{PCM}}$

        \State $x_{t_b}\gets
        \mathrm{ODEStep}
        \left(
        F_{\phi},
        x_{t_a},
        t_a\rightarrow t_b,c
        \right)$

        \State $x_{\tau_1}^{\mathrm{tar}}\gets
        \mathrm{sg}\left[
        G_{\theta}
        \left(
        x_{t_b},t_b\rightarrow\tau_1,c
        \right)
        \right]$

        \State $x_{\tau_1}^{\mathrm{pred}}\gets
        G_{\theta}
        \left(
        x_{t_a},t_a\rightarrow\tau_1,c
        \right)$

        \State $\mathcal{L}_{\mathrm{PCM}}\gets
        \left\|
        x_{\tau_1}^{\mathrm{pred}}
        -x_{\tau_1}^{\mathrm{tar}}
        \right\|_2^2$

        \Statex \textbf{// Update student}

        \State $\mathcal{L}_{G}\gets
        \lambda_{\mathrm{DMD}}\mathcal{L}_{\mathrm{DMD}}
        +
        \lambda_{\mathrm{PCM}}\mathcal{L}_{\mathrm{PCM}}$

        \State $\theta\gets
        \theta-\eta_{\theta}
        \nabla_{\theta}\mathcal{L}_{G}$

    \EndIf
\EndFor

\State \Return $G_{\theta}$
\end{algorithmic}
\end{algorithm}

\end{document}